\documentclass[lettersize,journal]{IEEEtran}
\usepackage{amsmath,amsfonts}
\usepackage{algorithmic}
\usepackage{array}
\usepackage[caption=false,font=normalsize,labelfont=sf,textfont=sf]{subfig}
\usepackage{textcomp}
\usepackage{stfloats}
\usepackage{url}
\usepackage{verbatim}

\usepackage{caption}
\usepackage[table]{xcolor}
\usepackage{array}
\usepackage{booktabs}
\usepackage{multirow}
\usepackage{placeins}
\usepackage{threeparttable}
\usepackage{booktabs}
\usepackage{multirow}
\usepackage{makecell}
\usepackage{graphicx}
\usepackage{tabularx}

\newlength{\biosep}
\usepackage[numbers,sort&compress]{natbib}  
\def\BibTeX{{\rm B\kern-.05em{\sc i\kern-.025em b}\kern-.08em
    T\kern-.1667em\lower.7ex\hbox{E}\kern-.125emX}}
\usepackage{balance}
\usepackage{hyperref}
\begin{document}
\bibliographystyle{plain}
\title{DRGBT-1K: A Large-scale High-quality Benchmark for Dynamic RGBT Tracking
}
\author{Zhaodong Ding, Chenglong Li, Senior Member IEEE, Zeyu Ding, Futian Wang, and Jin Tang
\thanks{This work was supported in
the National Natural Science Foundation of China under Grant No.62572004 and No.62376004.(\textit{Corresponding author: Chenglong Li, Futian Wang.})}
\thanks{Zhaodong Ding is with School of Artificial Intelligence, State Key Laboratory of Opto-Electronic Information Acquisition and Protection Technology and Anhui Provincial Key Laboratory of Security Artificial Intelligence,
Anhui University, Hefei 230601, China. 
(e-mail: zhaodongding\_ah@163.com; lcl1314@foxmail.com)}
\thanks{Chenglong Li is with State Key Laboratory of Opto-Electronic Information Acquisition and Protection Technology and Anhui Provincial Key Laboratory of Security Artificial Intelligence,
Anhui University, Hefei 230601, China. 
(e-mail: lcl1314@foxmail.com)}
\thanks{Zeyu Ding and Futian Wang are with, School of Computer Science, Anhui University, Hefei 230601, China. 
(e-mail: e124302149@stu.ahu.edu.cn; wft@ahu.edu.cn)}
\thanks{Jin Tang is with Anhui Provincial Key Laboratory of Multimodal Cognitive Computation, School of Computer Science and Technology and Anhui University, Hefei 230601, China. (e-mail: tangjin@ahu.edu.cn)
}}

\markboth{Journal of \LaTeX\ Class Files,~Vol.~18, No.~9, September~2020}%
{How to Use the IEEEtran \LaTeX \ Templates}

\maketitle

\begin{abstract}
Dynamic RGBT (DRGBT) tracking aims to continuously localize a target when the available sensing modalities and observation platforms vary over time. Compared with conventional RGBT tracking with fixed RGBT inputs and a fixed observation platform, DRGBT tracking is more consistent with real-world collaborative perception systems, where targets may be observed by heterogeneous sensors from different viewpoints. However, existing benchmarks are still insufficient for systematically evaluating tracker robustness under real dynamic modality variations and cross-platform transitions. To address this limitation, we make the following contributions. 
1) We construct DRGBT-1K, a large-scale high-quality benchmark for DRGBT tracking. It contains 1,045 sequences captured entirely in real-world scenarios and 795K RGBT frame pairs collected using UAVs and handheld RGBT devices, encompassing diverse real-world scenes, pronounced viewpoint changes, modality variations, and target appearance discontinuities.
2) We provide comprehensive annotations for fine-grained evaluation, including dense bounding boxes, target category labels, challenge attributes, frame-level modality labels and platform labels. DRGBT-1K covers 24 target categories, more than 15 scene types and 15 challenge attributes. 
3) We establish a comprehensive benchmark by evaluating 20 representative multimodal tracking methods, including conventional RGBT trackers, modality-missing RGBT trackers, and DRGBT trackers under a unified evaluation protocol. 
4) We release an unaligned version of DRGBT-1K and derive UGVT-1K to support broader research on unaligned multimodal tracking and UAV-ground collaborative tracking. 5) We develop an online evaluation platform for DRGBT-1K and provide a leaderboard that collects all methods evaluated on this benchmark.
These dataset resources and the leaderboard link are available at https://github.com/zhaodongAH/DRGBT-1K.
%


%
\end{abstract}

\begin{IEEEkeywords}
DRGBT tracking, Benchmark dataset, Large-scale, High-diversity.
\end{IEEEkeywords}

\begin{figure}[!ht]
\centering
\includegraphics[width=3.6in]{ 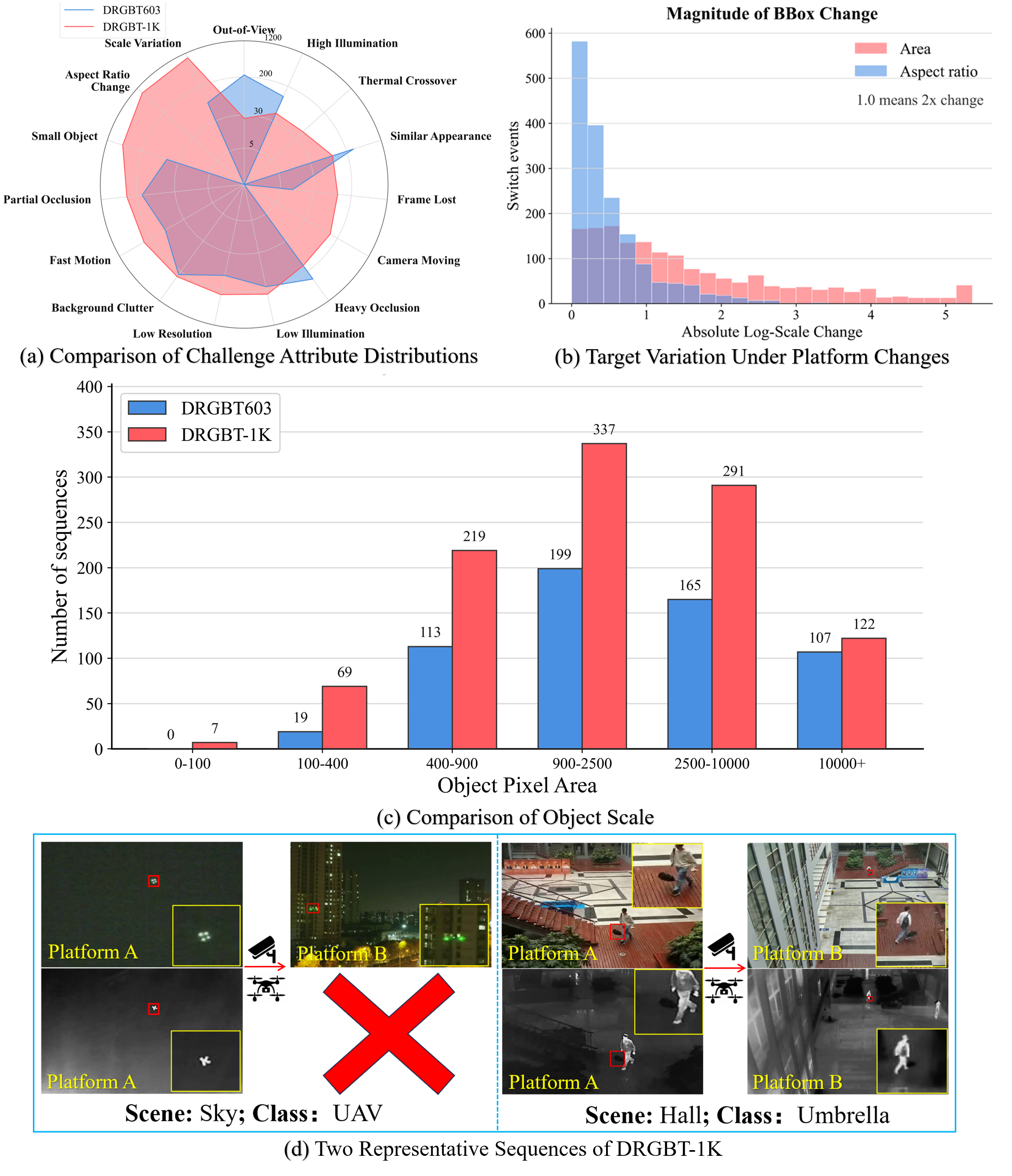}
\caption{Statistics and representative examples of DRGBT-1K.
(a) Challenge attribute distribution comparison between DRGBT-1K and DRGBT603.
(b) Target variations under platform switches, measured by absolute log-scale changes in bounding-box area and aspect ratio.
(c) Object pixel area distribution comparison between DRGBT-1K and DRGBT603.
(d) Representative sequences from DRGBT-1K, showing real-world cross-platform tracking scenarios with pronounced viewpoint changes, modality differences, scale variations, and appearance discontinuities.
}
\label{fig_DRGBT-1K}
\end{figure}

\section{Introduction}
RGBT tracking has attracted increasing attention due to its ability to exploit the complementary information of visible-light (RGB) and thermal-infrared (TIR) modalities for robust object tracking under challenging conditions~\cite{QAT2023,ding2025quality,feng2024rgbt}. 
Existing RGBT benchmarks and trackers usually assume that RGB and TIR data are captured from a single platform with synchronized sensors~\cite{lasher,rgbt210,rgbt234,VTUAV}. 
In this configuration, both the available modalities and the observation platform are held fixed throughout the sequence.
However, practical tracking scenarios are often far more complex than a fixed observational setup. To maintain continuous tracking in dynamic environments, a target may need to be handed over among different platforms, such as UAVs, ground-based cameras, handheld devices, and static surveillance cameras.
During this process, the accessible sensing modalities may also change over time, leading to sequences that contain RGB-only, TIR-only, or RGBT observations at different timestamps. 
This motivates the formulation of dynamic RGBT (DRGBT) tracking, where the tracker is expected to continuously estimate the target state despite dynamic changes in both sensing modalities and observation platforms. Such a task is more consistent with real deployment scenarios, but it also introduces pronounced variations in target appearance, viewpoint, scale and position, making it substantially more challenging than conventional RGBT tracking with fixed modalities and a fixed platform.

Although the first DRGBT tracking dataset, DRGBT603~\cite{DRGBT603}, was introduced to advance the development of DRGBT tracking, its sequence scale and scenario diversity remain limited. Only a subset of the sequences in DRGBT603 is constructed from real-captured data, which limits its ability to characterize real-world cross-platform tracking scenarios. To address this limitation, we construct a new large-scale DRGBT tracking benchmark, termed DRGBT-1K, which is composed entirely of real-captured sequences. DRGBT-1K contains 1,045 sequences and 795K RGBT image pairs, covering 24 target categories and 15 challenge attributes.
As shown in Fig.~\ref{fig_DRGBT-1K}(a), DRGBT-1K covers a wide range of challenge attributes and exhibits more diverse attribute distributions than DRGBT603, indicating its ability to characterize complex real-world tracking conditions. Fig.~\ref{fig_DRGBT-1K}(b) further shows that platform switches often lead to evident target variations, including changes in bounding-box area and aspect ratio. In addition, the object pixel area distribution in Fig.~\ref{fig_DRGBT-1K}(c) demonstrates that DRGBT-1K contains targets with more diverse scales, ranging from small objects to large targets.
We also present two representative sequences from DRGBT-1K in Fig.~\ref{fig_DRGBT-1K}(d). These examples show that the same target can undergo pronounced viewpoint changes, modality differences, scale variations, and appearance discontinuities during cross-platform transitions. These characteristics make DRGBT-1K a more realistic evaluation platform that closely reflects real-world dynamic tracking scenarios and promotes the development of more robust multimodal trackers.

Moreover, DRGBT-1K provides detailed annotations and comprehensive statistical analysis to further characterize these dynamic factors at the dataset level.
Specifically, each frame pair is spatially aligned and manually annotated with a target bounding box, which enables accurate evaluation across different modalities and observation platforms. We further analyze the dataset from multiple perspectives, including target categories, challenge attributes, target size distribution, sequence length distribution, modality variation, and platform transition. 
These annotations and statistics not only reveal the diversity and difficulty of DRGBT-1K, but also provide a solid basis for fine-grained analysis and systematic evaluation of DRGBT trackers.
To establish a comprehensive benchmark, we evaluate a wide range of representative RGB, RGBT, and DRGBT trackers under a unified evaluation protocol.
Experimental results show that existing trackers still suffer from significant performance degradation under real dynamic modality and platform variations.
In particular, trackers designed for conventional RGBT tracking are often sensitive to modality absence, viewpoint discontinuity, and abrupt target position changes.

The main contributions of this work are summarized as follows:
\begin{itemize}
    \item We construct DRGBT-1K, a large-scale high-quality benchmark for dynamic RGBT tracking. It contains 1,045 fully real-captured sequences and 795K RGBT frame pairs collected by UAVs and handheld RGBT devices, providing a realistic evaluation platform for cross-platform and modality-variant tracking.

    \item We provide comprehensive annotations to support fine-grained DRGBT tracking analysis. DRGBT-1K includes dense bounding boxes, target category labels, challenge attributes, frame-level modality labels, and platform labels, covering 24 target categories, more than 15 scene types, and 15 challenge attributes.

    \item We establish a systematic benchmark by evaluating 20 representative multimodal tracking methods under a unified protocol. The results reveal that existing trackers still face severe challenges under real modality variations and platform transitions, highlighting the necessity of developing robust DRGBT tracking methods.

    \item We release an unaligned version of DRGBT-1K and construct UGVT-1K derived from DRGBT-1K to support broader research on unaligned multimodal tracking and UAV-ground collaborative tracking. These resources provide additional foundations for studying realistic cross-modal and cross-platform visual tracking.

    \item We develop an online evaluation platform and leaderboard for DRGBT-1K to facilitate fair comparison and community development. The leaderboard provides \href{https://dongdong2061.github.io/DRGBT-1K-Leaderboard/}{links} to the corresponding code, model weights, and tracking results, while the test annotations are kept private to prevent overfitting or manual tuning based on test-set ground truth.
\end{itemize}

\section{Related Work}
\subsection{Datasets for RGBT Tracking}
\noindent \textbf{RGBT Tracking Datasets.}
In recent years, with the rapid development of deep learning, RGBT tracking has become one of the fundamental tasks in computer vision~\cite{feng2024rgbt}. 
In parallel with the development of RGBT tracking algorithms, a series of RGBT tracking benchmarks have been constructed to provide standardized evaluation platforms for this task. 
Early datasets mainly focus on paired RGBT sequences captured by fixed or relatively constrained imaging systems, enabling the community to evaluate the effectiveness of multimodal representation learning and RGBT fusion under challenging scenarios such as illumination variation, occlusion, background clutter, and thermal crossover~\cite{gtot,rgbt210,rgbt234}.
For example, GTOT~\cite{gtot} is one of the earliest benchmarks specifically designed for grayscale-thermal tracking, which contains 50 paired sequences collected under different scenarios and provides annotations for several representative challenges, thereby offering an initial standardized platform for evaluating the complementarity between visible and thermal modalities.
Subsequently, RGBT210~\cite{rgbt210} enlarges the benchmark scale to 210 paired RGBT sequences and provides more challenging attributes for comprehensive evaluation.
RGBT234~\cite{rgbt234} further extends RGBT210 to 234 sequences with more accurate annotations, richer challenging factors, and more comprehensive evaluation protocols, which makes it a widely used benchmark for evaluating conventional RGBT trackers.
With the increasing demand for training data in deep RGBT tracking, LasHeR~\cite{lasher} significantly expands the data scale and diversity by providing 1,224 aligned RGBT sequence pairs with more than 730K frame pairs, covering diverse object categories, camera viewpoints, scenes, environmental conditions, and 19 challenge attributes.
VTUAV~\cite{VTUAV} further introduces a large-scale visible-thermal UAV tracking benchmark, which contains 500 high-resolution UAV sequences with about 1.7M RGBT frame pairs.

\noindent \textbf{DRGBT Tracking Datasets.}
Despite the remarkable progress of the above RGBT benchmarks, they generally assume that RGB and TIR data are synchronously available and captured from the same or relatively stable platform throughout a sequence~\cite{IPT,catpp,sttrack}.
Such a setting is suitable for evaluating conventional RGBT fusion strategies, but it cannot fully reflect dynamic real-world scenarios, where the target may be observed by different platforms or only a subset of modalities is available at certain moments.
To explore this problem, DRGBT603~\cite{DRGBT603} introduces the DRGBT tracking task, where modality changes and platform switches are explicitly considered.
It contains 603 sequences with about 1.49M frame pairs, and provides frame-level modality labels, platform-switch labels, and sequence-level challenge annotations.
By incorporating dynamic modality variations and cross-platform observations, DRGBT603 provides an important benchmark for evaluating tracker robustness under more practical and challenging conditions.
Nevertheless, DRGBT tracking is still at an early stage.
Existing data remain limited in the scale and diversity of fully real-captured dynamic sequences, and current benchmarks are still insufficient to comprehensively evaluate robust tracking under frequent viewpoint changes, platform transitions, modality variations, and long-term target appearance changes.
To this end, we construct a new dataset, termed DRGBT-1K, which is collected using UAVs and ground cameras equipped with RGB and TIR sensors. DRGBT-1K contains 1,045 sequences with a total of 795K frame pairs.
Compared with the previous DRGBT tracking dataset DRGBT603, our dataset has three major differences. First, all sequences in DRGBT-1K are newly collected using UAVs and ground cameras equipped with RGB and TIR sensors. Second, DRGBT-1K contains real-world data with richer scenes and more diverse challenge categories. Finally, we provide multi-view RGBT tracking data as well as an unaligned version of DRGBT-1K, offering more comprehensive support for realistic DRGBT tracking.
A more detailed comparison between DRGBT-1K and other RGBT tracking datasets is presented in Table~\ref{tab:dataset_comparison}.

\begin{table*}[ht]
\caption{Comprehensive comparison of RGBT/DRGBT tracking datasets. The $\checkmark$ and $\times$ symbols indicate presence or absence, respectively.}
\centering
\small
\setlength{\tabcolsep}{2.3pt}
\renewcommand{\arraystretch}{1.25}
\resizebox{\textwidth}{!}{
\begin{tabular}{l|c|c|c|c|c|cc|c|c|c|c}
\toprule
\multirow{2}{*}{Dataset} 
& \multirow{2}{*}{Pub. Info.} 
& \multirow{2}{*}{Task Type}
& \multirow{2}{*}{View Num.}
& \multirow{2}{*}{\makecell{Sequence\\Num.}}
& \multirow{2}{*}{\makecell{Total\\Frames}}
& \multicolumn{2}{c|}{Object Info.} 
& \multirow{2}{*}{\makecell{Dynamic\\Modality}} 
& \multirow{2}{*}{\makecell{Cross\\Platform}} 
& \multirow{2}{*}{\makecell{Real Cross-platform\\Seq. Num.}}
& \multirow{2}{*}{\makecell{Synthetic Cross-platform\\Seq. Num.}} \\
& & & & & & Classes & Attr. & & & & \\
\midrule
GTOT~\cite{gtot}        & TIP 2017   & RGBT  & 1 & 50        & 7.8K   & 9  & 7  & $\times$     & $\times$     & 0             & 0 \\
RGBT210~\cite{rgbt210}  & CVPR 2018  & RGBT  & 1 & 210       & 104.7K & 22 & 12 & $\times$     & $\times$     & 0             & 0 \\
RGBT234~\cite{rgbt234}  & TPAMI 2019 & RGBT  & 1 & 234       & 116.7K & 22 & 12 & $\times$     & $\times$     & 0             & 0 \\
LasHeR~\cite{lasher}    & TIP 2021   & RGBT  & 1 & 1224      & 734.8K & 32 & 19 & $\times$     & $\times$     & 0             & 0 \\
VTUAV~\cite{VTUAV}      & CVPR 2022  & RGBT  & 1 & 500       & 1.7M   & 13 & 13 & $\times$     & $\times$     & 0             & 0 \\
\hline
DRGBT603~\cite{DRGBT603} & TIP 2026  & DRGBT & 2 & 603       & 1.49M  & 29 & 12 & $\checkmark$ & $\checkmark$ & 203           & 400 \\
DRGBT-1K (Ours)          & $-$       & DRGBT & 2 & 1045  & 795K   & 24 & 15 & $\checkmark$ & $\checkmark$ & \textbf{1045} & \textbf{0} \\
\bottomrule
\end{tabular}
}
\label{tab:dataset_comparison}
\end{table*}

\subsection{Methods for RGBT Tracking}
\noindent \textbf{RGBT Tracking Methods.}
RGBT tracking has attracted increasing attention in recent years, owing to its potential to achieve robust all-weather and day-and-night target tracking by exploiting the complementary properties of RGB and TIR modalities~\cite{ding2025quality,QAT2023,feng2024rgbt}.
Existing RGBT tracking methods mainly focus on effective multimodal fusion and spatio-temporal information modeling to enhance target representation and localization robustness in complex scenarios.
For example, Li et al.~\cite{tbsi} first introduce a cross-attention mechanism that uses the fused template as a bridge to facilitate cross-modal information interaction, thereby achieving effective multimodal fusion.
Subsequently, Ding et al.~\cite{ding2025quality} further exploit the fused template and cross-attention mechanism to design a token-level modality quality-aware module, enabling dynamic multimodal fusion according to modality reliability.
To address the limitation of fixed multimodal fusion strategies, Lu et al.~\cite{AFTER} propose a routing-driven dynamic attention module that adaptively adjusts the fusion structure according to the input.
In addition, spatio-temporal information modeling is also essential for RGBT tracking, as it enables trackers to exploit historical target states to handle challenges such as occlusion, deformation, and appearance variations~\cite{TATrack,ding2025template,ragtrack,sttrack}.
For example, Wang et al.~\cite{TATrack} propose a temporally adaptive tracking framework with a spatio-temporal two-stream architecture, where online template updating is employed to capture temporal information, while multimodal feature extraction and cross-modal interaction are jointly performed.
Li et al.~\cite{ragtrack} exploit a multimodal large language model to extract textual semantic cues and introduce retrieval-augmented generation to perform temporal semantic reasoning, thereby improving tracking stability.

\noindent \textbf{DRGBT Tracking Methods.}
Despite the remarkable progress of existing RGBT tracking methods, most of them are built upon the assumption of a single observation platform with available RGB and TIR modalities, which limits their applicability to cross-platform continuous tracking in real-world scenarios.
To overcome this limitation, Ding et al.~\cite{DRGBT603} introduce a causality-based modality- and platform-invariant representation learning framework to extract robust target representations against dynamic modality variations and platform switches.
However, research on DRGBT tracking methods is still in its early stage, and only limited efforts have been made to explicitly handle dynamic modality availability~\cite{IPT,TMKD} and cross-platform target re-localization~\cite{huang2024rtracker}.
Existing methods still face difficulties when target appearance changes drastically across platforms or when abrupt viewpoint and position shifts occur during platform transitions.
Moreover, the development of DRGBT trackers is largely constrained by the lack of large-scale real-captured benchmarks with diverse dynamic scenarios.
Therefore, constructing a more comprehensive DRGBT dataset is essential for evaluating existing trackers and promoting the development of robust cross-platform multimodal tracking methods.


\begin{figure*}[!ht]
\centering
\includegraphics[width=6.6 in]{ 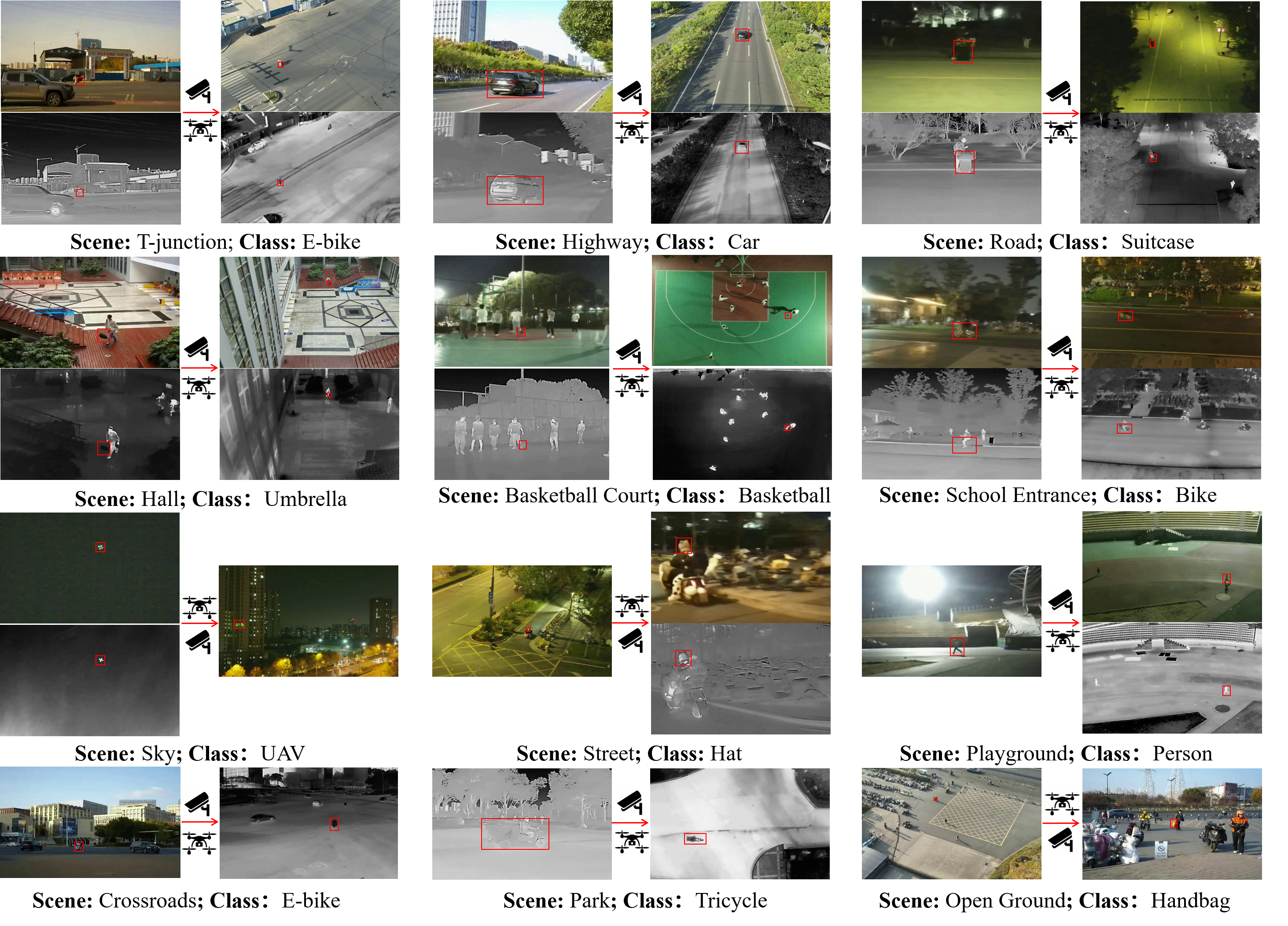}
\caption{Example frame pairs in DRGBT-1K. We present example sequences from different scenes and categories to demonstrate the diversity of DRGBT-1K.
}
\label{data_scene}
\end{figure*}

\section{DRGBT-1K Benchmark}
In this section, we introduce the proposed DRGBT-1K benchmark in detail. 
Different from conventional RGBT tracking datasets that are usually captured from a single observation platform with fixed RGBT inputs, DRGBT-1K is designed to support DRGBT tracking under more realistic cross-platform and modality-variant scenarios.
DRGBT-1K contains 1,045 fully real-captured sequences with a total of 795K RGBT frame pairs, covering 24 target categories and 15 challenge attributes.
Compared with previous RGBT and DRGBT benchmarks, DRGBT-1K provides larger-scale real dynamic sequences, richer real-world scenarios, and more diverse cross-platform observations.
In the following, we describe the data acquisition process, annotation protocol, dataset statistics, and evaluation protocols of DRGBT-1K.

\subsection{Dataset Collection and Alignment}
Existing DRGBT tracking datasets have two major limitations. First, part of the data is synthesized by processing existing RGBT tracking sequences. 
Although such data can simulate modality variations and abrupt target position shifts to some extent, it is difficult to faithfully characterize target appearance changes across different platforms and viewpoints. Second, although existing real-captured sequences are collected using multiple devices, their scale remains limited and the viewpoint variations are relatively small, making them insufficient to meet the diversity requirements of real-world cross-platform continuous tracking.
To this end, we employ multiple acquisition devices to collect data under diverse observation conditions.
Specifically, DRGBT-1K is collected using UAVs and ground cameras equipped with RGB and TIR sensors, enabling the same target to be observed from different platforms and viewpoints over time.
This acquisition setting introduces substantial appearance variations, viewpoint changes, scale changes, and modality differences, making DRGBT-1K more consistent with practical continuous tracking scenarios.
After data collection, we first perform temporal and spatial alignment for the multi-view sequences. Since different devices may have discrepancies in starting time and frame order during collaborative recording, temporal alignment is required to ensure that the target trajectory remains continuous and realistic after cross-platform switching. Specifically, we design a script to manually assist in cropping the sequences captured by two platforms, so that they are consistent in temporal range and sequence length.
Through this process, we obtain synchronized sequences of the same target from different viewpoints, which provides the basis for constructing cross-platform continuous tracking sequences.
Subsequently, we perform spatial alignment for the sequences from different viewpoints. Since the positions of the acquisition devices are not fixed, RGB and TIR images in different sequences exhibit varying degrees of spatial deviations, making it infeasible to use unified alignment parameters. Therefore, manually assisted calibration and registration are required. To this end, we design a new spatial alignment tool to align RGB and TIR images for each viewpoint sequence. The temporal alignment script and spatial alignment tool will be released together with the DRGBT-1K dataset to facilitate future research.
\subsection{Annotation}
In DRGBT-1K, we provide multiple types of high-quality annotations, including target bounding boxes, challenge attributes, target categories, and modality labels, to support comprehensive training and evaluation of DRGBT trackers. 
Representative examples are shown in Fig.~\ref{fig_DRGBT-1K}(b) and Fig.~\ref{data_scene}.

\begin{itemize}
    \item \textit{Bounding boxes.} 
    In DRGBT-1K, we carefully annotate targets from different viewpoints. Since existing annotation tools can hardly support synchronized annotation of the same target across multi-view sequences, we design a dual-view collaborative annotation tool. This tool uses a unified timestamp to simultaneously control two viewpoint sequences that have been temporally and spatially aligned, allowing annotators to label the same target across different views at the same time step. In this way, the consistency of cross-view target identity can be ensured. Finally, the entire dataset contains 799K annotated target bounding boxes.

    \item \textit{Attribute and category annotation.} 
    We provide sequence-level challenge attribute annotations to support attribute-based evaluation of tracking robustness. 
    Specifically, DRGBT-1K contains 15 challenge attributes that cover common difficulties in DRGBT tracking, such as occlusion, low illumination, background clutter, similar appearance, fast motion and scale variation.  
    In addition, we assign a category label to each sequence, resulting in 24 target categories in total, to characterize the target diversity of DRGBT-1K and facilitate future semantic-aware tracking research.

    \item \textit{Modality and platform labels.} 
    Unlike previous RGBT tracking datasets, we provide frame-level modality labels to indicate the available modality at each timestamp, including RGB-only, TIR-only, and RGBT observations. 
    In addition, platform labels are provided to indicate whether a platform transition occurs during tracking. 
    These annotations make it possible to evaluate tracker performance under dynamic modality changes and cross-platform switching, which are essential for realistic DRGBT tracking.
\end{itemize}

\begin{table}[ht]
\centering
\caption{List and description of 15 challenge attributes in DRGBT-1K.}
\label{tab:challenge_attributes}
\renewcommand{\arraystretch}{1.18}
\setlength{\tabcolsep}{3pt}
\begin{tabularx}{\columnwidth}{lX}
\toprule
\textbf{Attribute} & \textbf{Definition} \\
\midrule
HO  & \textit{Heavy Occlusion} -- the target is severely occluded. \\
PO  & \textit{Partial Occlusion} -- the target is slightly or partially occluded. \\
LI  & \textit{Low Illumination} -- the sequence is captured under low-light conditions, such as nighttime scenes. \\
LR  & \textit{Low Resolution} -- the target region is blurred or has low visual resolution. \\
BC  & \textit{Background Clutter} -- the background is complex and contains many interfering objects, such as trees or pedestrians. \\
HI  & \textit{High Illumination} -- strong illumination or glare appears in the scene, such as street lights or vehicle lights. \\
SA  & \textit{Similar Appearance} -- similar objects appear near the target, making the target difficult to distinguish. \\
FL  & \textit{Frame Lost} -- several consecutive frames are completely identical. \\
SO  & \textit{Small Object} -- the target is very small in the image. \\
FM  & \textit{Fast Motion} -- the target moves rapidly between adjacent frames. \\
OV  & \textit{Out-of-View} -- the target leaves the camera field of view and later reappears. \\
SV  & \textit{Scale Variation} -- the target undergoes significant scale changes. \\
CM  & \textit{Camera Moving} -- the camera has obvious motion or shake. \\
TC  & \textit{Thermal Crossover} -- the target has a similar temperature to surrounding objects or the background, making it difficult to distinguish in the thermal modality. \\
ARC & \textit{Aspect Ratio Change} -- the aspect ratio of the bounding box is outside the range $[0.5, 2]$. \\
\bottomrule
\end{tabularx}
\renewcommand{\arraystretch}{1.0}
\end{table}


\begin{figure*}[!ht]
\centering
\includegraphics[width=6.5 in]{ 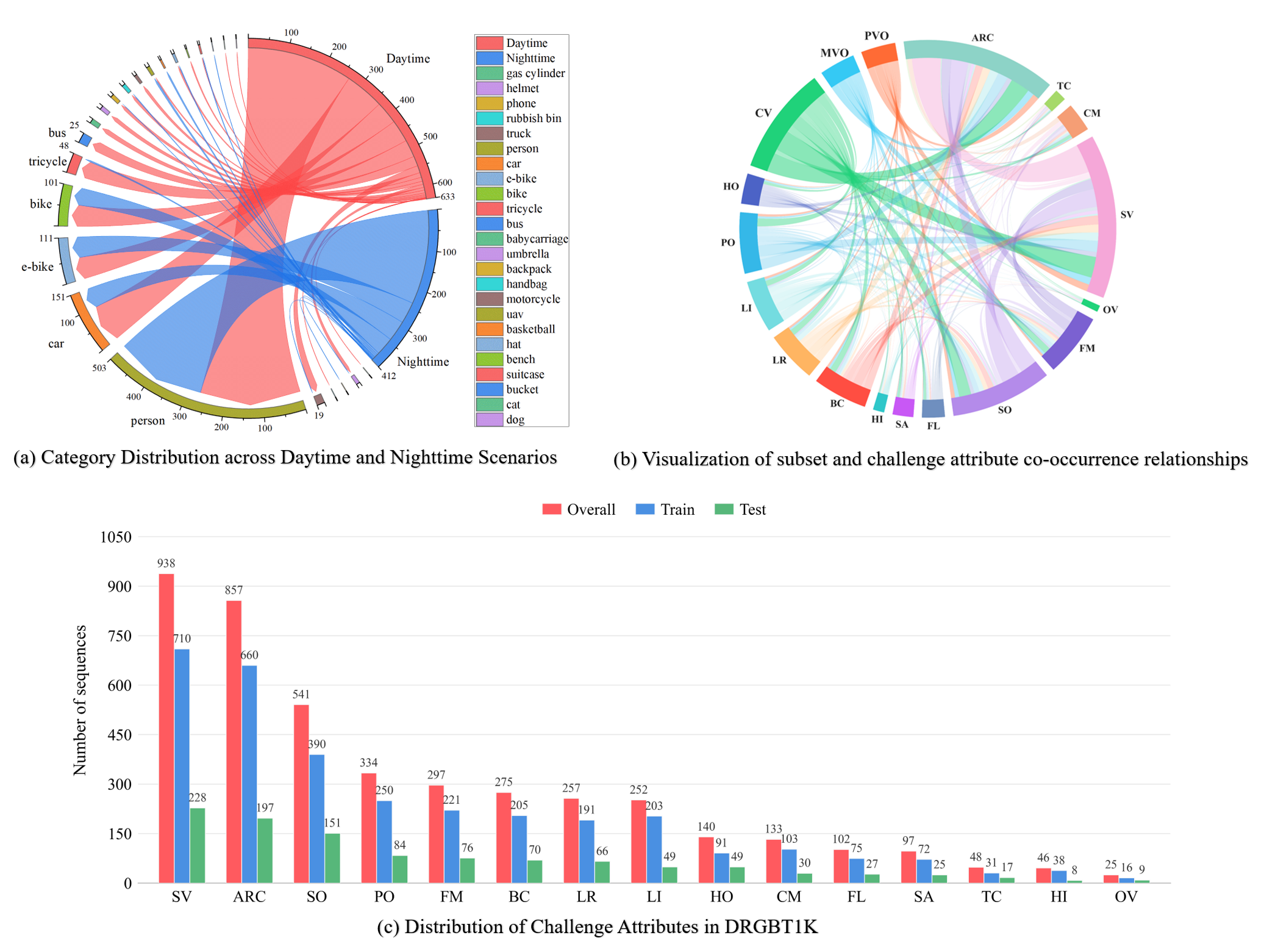}
\caption{Distribution of target categories, challenge attributes, and sequences under different modality and platform variations in DRGBT-1K.
}
\label{distribution}
\end{figure*}

\subsection{Data Statistics}
DRGBT-1K provides broad coverage of target categories, scene types, and viewpoint variations, reflecting the diversity of DRGBT tracking scenarios.
By adopting a multi-platform collaborative acquisition strategy and leveraging the flexibility of UAVs and ground cameras in viewpoints and motion ranges, we are able to collect real cross-platform RGBT data with substantial viewpoint variations, pronounced target appearance changes, diverse target categories, and a wide range of scene types.
To comprehensively demonstrate the advantages of DRGBT-1K, we analyze the diversity from the following aspects.

\noindent \textbf{Scene type.}
In visual tracking, scene complexity and diversity are critical factors for evaluating the robustness of tracking algorithms. To this end, we collect video data from a wide range of indoor and outdoor environments, covering more than 15 types of real scenes with distinct characteristics.
These scenes with varying levels of complexity pose different degrees of challenges to DRGBT trackers, which helps promote the development of more robust tracking algorithms. Some representative scenes from DRGBT-1K are shown in Fig.~\ref{data_scene}.

\noindent \textbf{Object category.}
DRGBT tracking aims to robustly localize targets of arbitrary categories across different platforms under dynamic viewpoint and modality variations. Therefore, broader coverage of target categories in the dataset is essential for comprehensively evaluating the generalization ability of trackers across diverse target types.
To this end, DRGBT-1K includes sequences from 24 target categories collected across different time periods, as illustrated in Fig.~\ref{distribution}(a).
It can be observed that the five most frequent target categories in DRGBT-1K are person, car, e-bike, bike, and tricycle, which also represent common target types in real scenarios.
In addition to these common categories, DRGBT-1K also contains many less frequent but challenging targets, such as drone, ball, and umbrella, which further enrich the category diversity of the dataset. This long-tailed category distribution better reflects practical tracking scenarios and encourages trackers to maintain robust performance on both common and rare targets.

\noindent \textbf{Challenge.}
Compared with existing DRGBT tracking datasets~\cite{DRGBT603}, DRGBT-1K provides a richer set of challenge attributes derived from real tracking scenarios.
Specifically, it contains 15 sequence-level attributes: Heavy Occlusion (HO), Partial Occlusion (PO), Low Illumination (LI), Low Resolution (LR), Background Clutter (BC), High Illumination (HI), Similar Appearance (SA), Frame Lost (FL), Small Object (SO), Fast Motion (FM), Out-of-View (OV), Scale Variation (SV), Camera Moving (CM), Thermal Crossover (TC), and Aspect Ratio Change (ARC). The detailed definitions of these attributes are summarized in Table~\ref{tab:challenge_attributes}.
We further present the distribution of sequences with different challenge attributes in DRGBT-1K, as shown in Fig.~\ref{distribution}(c).
It can be observed that, in DRGBT tracking scenarios, platform transitions usually introduce significant viewpoint variations, leading to substantial changes in target scale and aspect ratio during tracking.
As a result, SV and ARC are the two most frequent challenge attributes in DRGBT-1K.
SO ranks third, mainly because some targets appear at small scales from the UAV viewpoint.
These factors jointly increase the difficulty of target localization and state estimation, posing greater challenges to tracker robustness.

\begin{figure*}[!ht]
\centering
\includegraphics[width=6.6 in]{ 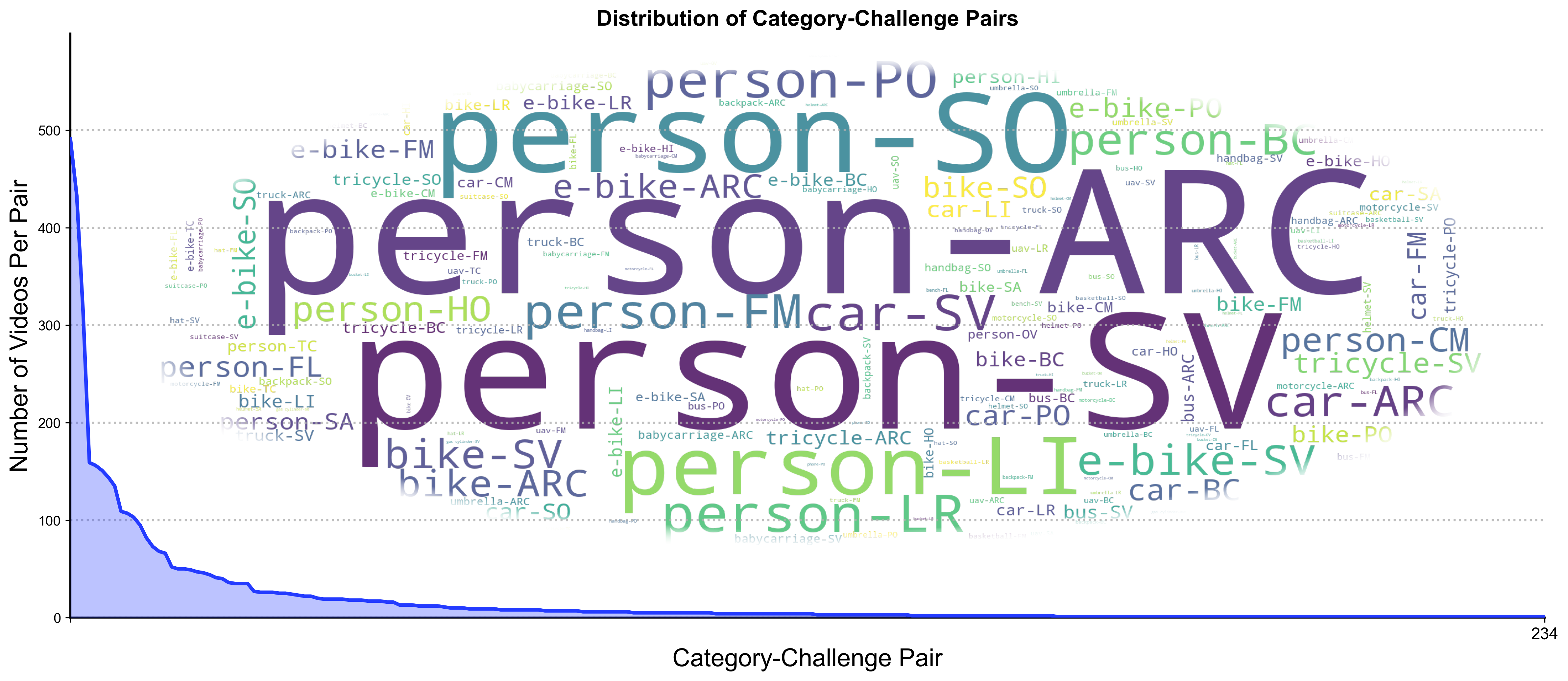}
\caption{Distribution and Word Cloud Visualization of Category-Challenge Pairs in DRGBT-1K.
}
\label{category-challenge}
\end{figure*}

We also present a word cloud of category-challenge pairs to further analyze the joint distribution between target categories and challenge attributes in complex dynamic scenarios.
As shown in Fig.~\ref{category-challenge}, person frequently co-occurs with several challenging attributes, such as SV, LR, PO, BC, and SO, indicating that pedestrian targets in DRGBT tracking are often affected by scale variation, low resolution, occlusion, background clutter, and small-object conditions.
Moreover, common traffic-related categories, including car, e-bike, bike, and tricycle, are also strongly associated with multiple challenge attributes, suggesting that DRGBT-1K covers complex interference factors across different target types in real cross-platform tracking scenarios.
This category-challenge joint distribution further demonstrates the rich diversity of DRGBT-1K in terms of both target categories and challenge combinations, providing a more comprehensive basis for evaluating the robustness and generalization ability of DRGBT trackers.

\begin{figure}[!ht]
\centering
\includegraphics[width=3.4in]{ 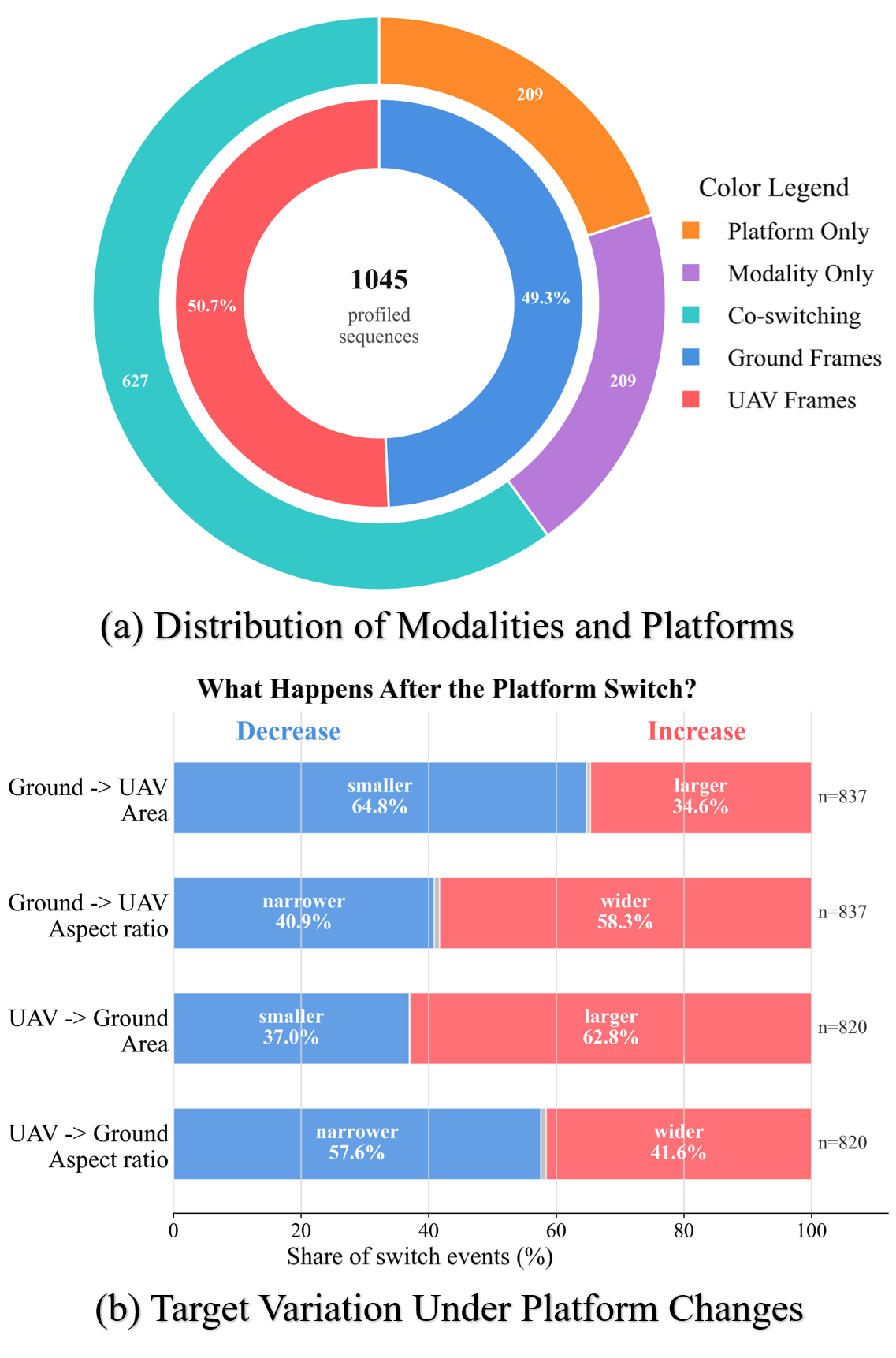}
\caption{Statistical analysis of modality, platform, and target variations after platform switches in DRGBT-1K.
(a) Distribution of dynamic variation patterns and platform viewpoints in DRGBT-1K.
(b) Target variations after platform switches in terms of bounding-box area and aspect ratio.
}
\label{variation}
\end{figure}

\begin{table}[!hb]
\centering
\caption{Statistics of DRGBT-1K splits. ``Len.'' denotes the number of frames.}
\label{tab:DRGBT-1K_split_stats}
\renewcommand{\arraystretch}{1.12}
\setlength{\tabcolsep}{6pt}
\begin{tabular}{lccc}
\hline
\textbf{Split} & \textbf{Seq.} & \textbf{Avg. Len.} & \textbf{Max Len.} \\
\hline
Train & 800 & 335.70 & 2733 \\
Test  & 245 & 525.65 & 4530 \\
\hline
\end{tabular}
\renewcommand{\arraystretch}{1.0}
\end{table}

\noindent \textbf{Modality and Platform Variations.}
To evaluate tracker performance under different dynamic variation conditions, we follow the same setting as DRGBT603~\cite{DRGBT603} and divide the sequences into three subsets: Modality Variation Only (MVO), Platform Variation Only (PVO), and Combined Variation (CV). Specifically, MVO contains sequences with only modality variation, PVO contains sequences with only platform variation, and CV contains sequences with both modality and platform variations.
As shown in Fig.~\ref{variation}(a), DRGBT-1K contains 209 MVO sequences, 209 PVO sequences, and 627 CV sequences. This distribution indicates that DRGBT-1K not only supports the evaluation of isolated modality or platform changes, but also emphasizes more challenging scenarios where modality variations and platform transitions occur simultaneously. In addition, the frame distribution between UAV and ground viewpoints is relatively balanced, accounting for 50.7\% and 49.3\%, respectively, which helps reduce viewpoint bias in cross-platform evaluation.
We further analyze target variations after platform switches, as shown in Fig.~\ref{variation}(b). The statistics show that transitions between ground and UAV viewpoints often lead to evident changes in target area and aspect ratio, reflecting substantial scale and shape variations caused by platform changes. Moreover, we present the coupling relationships between different subsets and challenge attributes in Fig.~\ref{distribution}(b). It can be observed that sequences from different subsets are often associated with multiple challenging factors simultaneously, further demonstrating the complexity and challenging nature of DRGBT-1K.

\subsection{Dataset Split and Evaluation Protocol}
\noindent \textbf{Dataset Split.}
DRGBT-1K contains 1,045 DRGBT tracking sequences, among which 800 sequences are used for training and the remaining 245 sequences are used for testing.
Table~\ref{tab:DRGBT-1K_split_stats} presents a statistical comparison between the training and testing sets of DRGBT-1K. During dataset splitting, we aim to maintain consistent distributions between the training and testing sets in terms of target categories, scene types, challenge attributes, and dynamic variation patterns, so that the testing results can objectively reflect the generalization ability of trackers.

\noindent \textbf{Evaluation Protocol.}
In our experiments, all trackers are evaluated under the one-pass evaluation (OPE) protocol. We adopt three standard metrics, including Precision Rate (PR), Normalized Precision Rate (NPR), and Success Rate (SR), to comprehensively assess tracking performance. 
Specifically, Precision Rate (PR) measures the percentage of frames whose center location error is within 20 pixels. Normalized Precision Rate (NPR) normalizes the center location error to reduce the influence of image resolution and target size~\cite{trackingnet}. Success Rate (SR) evaluates the overlap between the predicted and ground-truth bounding boxes, with the area under the success curve reported as the representative score.

\begin{figure}[!ht]
\centering
\includegraphics[width=3.5 in]{ 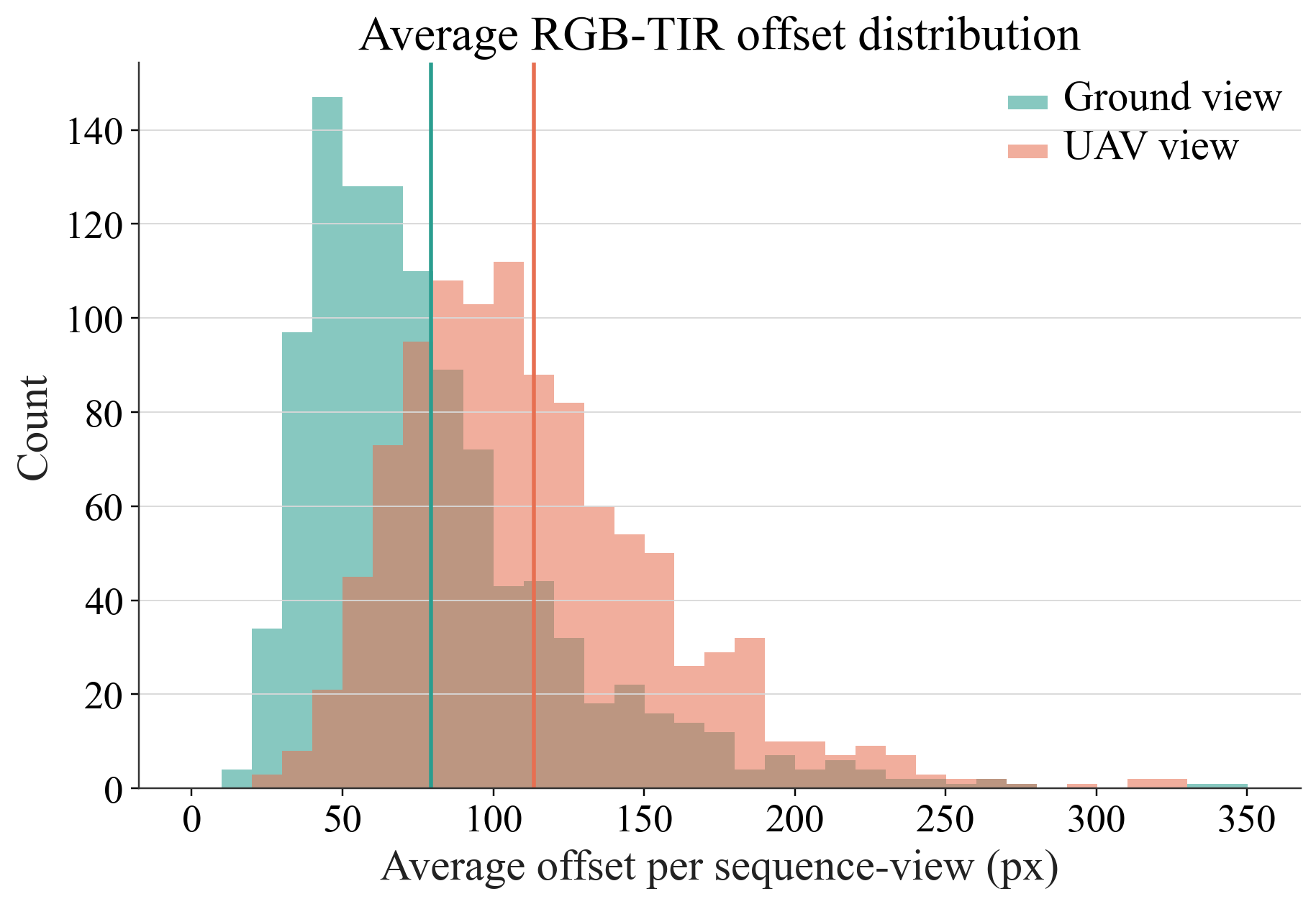}
\caption{Comparison of average modality offsets between UAV and ground-camera viewpoints.
}
\label{unaligned}
\end{figure}

\subsection{Unaligned DRGBT-1K}
In practical applications, due to differences in device structures, sensor mounting positions, and imaging mechanisms, the data captured by RGB and TIR sensors are usually unaligned or weakly aligned. 
However, accurate alignment between RGB and TIR modalities is often time-consuming and labor-intensive, which limits the practical deployment of DRGBT tracking algorithms in real scenarios. 
Although some methods~\cite{jin2026progressive,xiao2026unaligned} have begun to explore unaligned RGBT tracking, this research direction is still in its early stage. To further promote the real-world application of DRGBT tracking, we also release an unaligned version of DRGBT-1K, with several example sequences shown in Fig.~\ref{unaligned}.

To construct the unaligned version of DRGBT-1K, we preserve the original sequence structure and keep the RGB modality unchanged, while introducing controllable spatial perturbations only to the TIR modality. For each sequence and each viewpoint, we first load the original synchronized annotations and use the image diagonal length $D$ as the scale normalization reference. A frame-wise RGB–TIR offset vector is then generated.
Specifically, given the target center $c_t$ and the image center $o$ at frame $t$, we compute the spatial deviation $\psi_t=|c_t-o|_2$ and combine it with a rank-normalized temporal response to determine the final offset magnitude. The magnitude is constrained within $[0.01D, 0.17D]$, while the base direction is defined from the target center toward the image center. In addition, a sequence-level initial modality bias $\rho$ is randomly sampled within $[2\%,17\%]D$ with a random direction, simulating inherent cross-sensor misalignment across different sequences.
To further enhance realism, the generated offset sequence incorporates multiple temporal patterns, including synchronized, opposite, and delayed modes, as well as smooth, normal, medium-change, and impulse-change variations. The resulting offsets are applied to translate the TIR images accordingly. Moreover, with a certain probability, a center-based scale transformation is applied to the TIR modality to simulate cross-modal field-of-view variations.
The modality offsets between RGB and TIR under different viewpoints are illustrated in Fig.~\ref{unaligned}. Finally, the corresponding TIR bounding boxes are transformed using the same geometric parameters and clipped to image boundaries.

\begin{figure}[!ht]
\centering
\includegraphics[width=3.5 in]{ 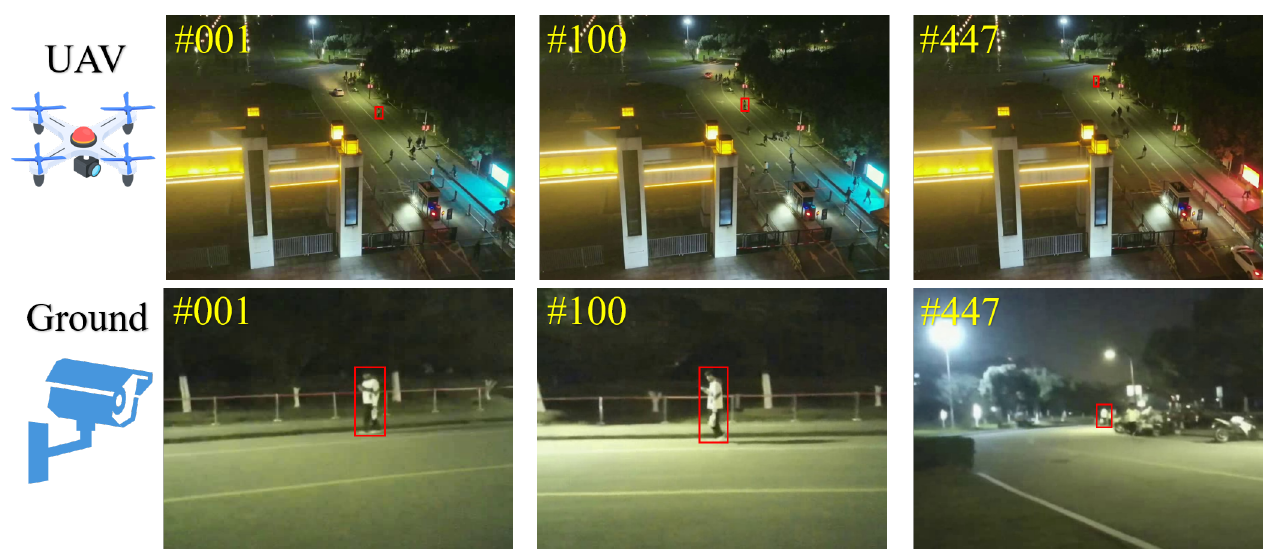}
\caption{A representative tracking example from UGVT-1K.
}
\label{ugvt1k}
\end{figure}

\subsection{UGVT-1K}

UAV-ground visual tracking has attracted increasing attention from the research community~\cite{sun2023uav,xu2026vl}, and has important applications in cross-view perception and real-world surveillance scenarios.
To support research in this direction, and considering that DRGBT-1K provides temporally aligned multi-view data with bounding box annotations, enabling synchronized tracking sequences across multiple viewpoints with rich category diversity, viewpoint variations, and real-world scene complexity.
Based on this foundation, we further construct a UAV-ground visual tracking benchmark by leveraging the DRGBT-1K dataset. Specifically, since UAV-ground RGB tracking does not require the TIR modality and certain redundant annotations, we remove them to obtain a new dataset, UGVT-1K.
Fig.~\ref{ugvt1k} shows a representative tracking sequence from UGVT-1K.
UGVT-1K preserves the original sequence structure and viewpoint diversity of DRGBT-1K, while focusing on UAV-ground RGB tracking scenarios. The dataset contains 1,045 sequences, including 800 for training and 245 for testing, following the same data splitting protocol as DRGBT-1K.

\begin{table}[!hb]
\centering
\caption{Comparison with state-of-the-art trackers in DRGBT-1K. The best and second-best results are highlighted in red and blue, respectively.}
\label{tab:sota_comparison}
\renewcommand{\arraystretch}{1.18}
\resizebox{\columnwidth}{!}{
\begin{tabular}{lcccc}
\hline
\textbf{Method} & \textbf{Source} & \textbf{PR} & \textbf{NPR} & \textbf{SR} \\
\hline
\multicolumn{5}{c}{\textit{RGBT Trackers}} \\
\hline
OSTrack~\cite{ye2022ostrack}   & ECCV'22   & 47.39 & 41.47 & 34.46 \\
TBSI~\cite{tbsi}               & CVPR'23   & 43.89 & 38.81 & 32.34 \\
TATrack~\cite{TATrack}         & AAAI'24   & 47.82 & 41.17 & 34.40 \\
BAT~\cite{BAT}                 & AAAI'24   & 45.84 & 41.08 & 33.86 \\
PURA~\cite{PURA}               & CVPR'24   & 48.32 & 40.73 & 33.98 \\
SDSTrack~\cite{SDSTrack}       & CVPR'24   & 40.66 & 33.64 & 28.23 \\
MMLoRAT~\cite{LoRAT}           & ECCV'24   & 47.76 & \textcolor{blue}{43.06} & 34.92 \\
CKD~\cite{CKD}                 & ACM MM'24 & 45.67 & 40.97 & 33.75 \\
AINet~\cite{ainet}             & AAAI'25   & 44.83 & 40.42 & 33.14 \\
STTrack~\cite{sttrack}         & AAAI'25   & 22.57 & 17.38 & 16.61 \\
CAFormer~\cite{CAFormer}       & AAAI'25   & 46.06 & 39.80 & 33.51 \\
FMTrack~\cite{xue2025fmtrack}  & TCSVT'25  & \textcolor{blue}{48.64} & 41.58 & 34.91 \\
QSTNet~\cite{ding2025quality}  & TIP'25    & 44.21 & 39.20 & 32.27 \\
MRTTrack~\cite{MRTTrack}       & PR'25     & 43.72 & 38.76 & 32.04 \\
UATrack~\cite{UATrack}         & IJCV'26   & 46.92 & 40.77 & 34.02 \\
GOLA~\cite{GOLA}               & CVPR'26   & \textcolor{red}{51.18} & \textcolor{red}{46.42} & \textcolor{red}{38.01} \\
\hline
\multicolumn{5}{c}{\textit{MMRGBT Trackers}} \\
\hline
IPT~\cite{IPT}                 & IJCV'25   & 46.52 & 39.89 & 33.46 \\
TMKD~\cite{TMKD}               & PR'26     & 48.06 & 42.95 & \textcolor{blue}{35.52} \\
SCDT~\cite{SCDT}               & CVPR'26   & 45.57 & 36.41 & 29.91 \\
\hline
\multicolumn{5}{c}{\textit{DRGBT Trackers}} \\
\hline
CMRL~\cite{DRGBT603}           & TIP'26    & 43.03 & 38.48 & 31.82 \\
\hline
\end{tabular}
}
\renewcommand{\arraystretch}{1.0}
\end{table}

\begin{table*}[t]
\centering
\caption{Attribute- and subset-based performance on the DRGBT-1K test set. Each cell reports PR/SR (\%). The best and second-best results under each attribute are highlighted in red and blue, respectively.}
\label{tab:challenge_pr_sr_transposed}
\renewcommand{\arraystretch}{1.12}
\setlength{\tabcolsep}{2.2pt}
\scriptsize

\resizebox{\textwidth}{!}{
\begin{tabular}{lcccccccccc}
\hline
\textbf{Attr.} & \textbf{OSTrack~\cite{ye2022ostrack}} & \textbf{TBSI~\cite{tbsi}} & \textbf{TATrack~\cite{TATrack}} & \textbf{BAT~\cite{BAT}} & \textbf{PURA~\cite{PURA}} & \textbf{SDSTrack~\cite{SDSTrack}} & \textbf{MMLoRAT~\cite{LoRAT}} & \textbf{CKD~\cite{CKD}} & \textbf{AINet~\cite{ainet}} & \textbf{STTrack~\cite{sttrack}} \\
\hline
HO  & \textcolor{red}{47.97}/\textcolor{blue}{34.64} & 39.38/30.55 & \textcolor{blue}{47.41}/34.02 & 42.73/32.60 & 44.72/31.56 & 33.00/25.14 & 42.33/30.90 & 43.36/32.80 & 39.42/31.08 & 13.41/8.73 \\
PO  & 39.01/32.05 & 36.61/30.00 & 39.05/31.84 & 36.74/30.64 & 36.63/29.95 & 32.79/26.40 & 38.22/30.60 & 37.05/30.55 & 36.26/30.08 & 7.79/8.62 \\
LI  & 34.72/29.56 & 35.62/30.07 & \textcolor{blue}{38.13}/30.91 & \textcolor{red}{38.17}/\textcolor{blue}{31.16} & 36.47/30.00 & 30.13/23.84 & 36.20/30.34 & 35.11/29.83 & 34.95/29.72 & 9.97/7.94 \\
LR  & 36.17/27.70 & 33.56/26.44 & 36.93/28.40 & 36.06/28.27 & 35.57/26.41 & 31.23/23.65 & 34.85/27.55 & 34.17/27.46 & 33.59/27.13 & 13.49/8.01 \\
BC  & \textcolor{blue}{44.04}/33.73 & 35.94/29.49 & \textcolor{red}{44.82}/\textcolor{red}{34.68} & 41.37/32.57 & 40.97/30.47 & 33.12/26.03 & 38.66/30.40 & 39.56/31.82 & 36.25/30.04 & 15.47/9.37 \\
HI  & 33.41/24.90 & 31.51/22.72 & 28.56/22.22 & 29.78/20.53 & 24.43/18.97 & 25.61/17.75 & 24.14/18.17 & 34.72/24.54 & 26.51/20.06 & 4.46/6.08 \\
SA  & 40.91/\textcolor{blue}{32.99} & 37.60/30.84 & 39.85/31.74 & 40.47/32.64 & 38.92/29.38 & 29.22/23.85 & 37.54/29.63 & 37.97/31.33 & 35.31/30.12 & 12.25/9.16 \\
FL  & 35.82/25.15 & 34.22/24.77 & 36.98/\textcolor{blue}{26.96} & 34.56/24.87 & 32.62/22.86 & 27.21/20.44 & 34.55/24.96 & 34.05/25.05 & 32.17/23.76 & 6.69/6.55 \\
SO  & 41.10/30.99 & 36.97/28.83 & \textcolor{red}{42.33}/\textcolor{blue}{31.75} & 38.21/29.49 & 38.40/27.96 & 32.29/24.31 & 38.84/29.13 & 38.22/29.53 & 35.90/28.35 & 14.88/8.41 \\
FM  & 38.39/32.59 & 37.32/32.06 & 39.95/33.64 & 40.33/34.22 & 38.17/31.84 & 31.42/26.67 & 36.70/31.83 & 37.27/32.24 & 36.05/31.54 & 5.79/9.19 \\
OV  & 16.78/15.26 & 14.25/13.48 & 16.06/14.45 & 17.90/14.61 & \textcolor{blue}{18.90}/\textcolor{blue}{15.78} & 12.41/11.62 & 16.76/14.98 & 16.81/14.24 & 15.40/13.80 & 7.27/7.05 \\
SV  & 40.48/32.36 & 36.97/30.34 & \textcolor{red}{41.62}/\textcolor{blue}{33.22} & 39.21/31.78 & 39.22/30.43 & 32.83/26.18 & 38.74/30.94 & 38.10/31.27 & 36.69/30.32 & 11.54/8.79 \\
CM  & 41.00/34.23 & 38.34/33.05 & 42.60/35.59 & \textcolor{blue}{44.45}/\textcolor{blue}{36.67} & 39.91/31.59 & 35.05/28.75 & 37.16/31.96 & 40.46/34.05 & 40.62/34.15 & 6.10/9.42 \\
TC  & 25.77/21.71 & 24.86/20.57 & 27.99/22.06 & 25.22/20.94 & 29.84/22.53 & 19.93/15.83 & 31.42/23.87 & 26.77/22.23 & 23.78/19.91 & 4.50/7.26 \\
ARC & 39.49/31.77 & 35.95/29.81 & \textcolor{red}{41.19}/\textcolor{blue}{32.96} & 37.74/30.94 & 38.07/29.83 & 32.04/25.62 & 37.79/30.37 & 37.11/30.62 & 35.80/29.73 & 11.09/8.56 \\
\hline
MVO & 77.25/54.24 & 65.01/48.22 & \textcolor{red}{78.72}/\textcolor{blue}{56.21} & 69.83/52.08 & 70.42/49.35 & 51.20/36.79 & 69.09/50.30 & 69.34/51.29 & 64.67/48.24 & 21.01/11.41 \\
PVO & 35.12/30.14 & 35.71/30.29 & 36.97/30.67 & 37.02/31.09 & 37.74/29.91 & 35.24/27.70 & 35.81/30.06 & 36.11/30.86 & 35.53/30.64 & 11.11/8.46 \\
CV  & 30.19/25.36 & 28.32/24.18 & 30.92/25.76 & 29.55/24.65 & 29.75/24.18 & 26.34/21.86 & 29.89/24.55 & 28.72/24.42 & 28.20/24.12 & 8.92/8.00 \\
\hline
\end{tabular}
}

\vspace{1mm}

\resizebox{\textwidth}{!}{
\begin{tabular}{lcccccccccc}
\hline
\textbf{Attr.} & \textbf{CAFormer~\cite{CAFormer}} & \textbf{IPT~\cite{IPT}} & \textbf{FMTrack~\cite{xue2025fmtrack}} & \textbf{QSTNet~\cite{ding2025quality}} & \textbf{UATrack~\cite{UATrack}} & \textbf{CMRL~\cite{DRGBT603}} & \textbf{MRTTrack~\cite{MRTTrack}} & \textbf{TMKD~\cite{TMKD}} & \textbf{SCDT~\cite{SCDT}} & \textbf{GOLA~\cite{GOLA}} \\
\hline
HO  & 39.64/30.46 & 37.03/27.82 & 40.95/30.98 & 38.79/29.71 & 46.84/33.91 & 46.71/34.56 & 40.02/30.04 & 46.10/\textcolor{red}{34.83} & 24.74/18.14 & 45.30/34.15 \\
PO  & 37.34/30.13 & 33.87/28.05 & 38.15/29.81 & 35.90/29.51 & 39.58/31.95 & \textcolor{blue}{41.13}/\textcolor{blue}{33.82} & 34.26/28.69 & 38.99/32.07 & 29.53/21.35 & \textcolor{red}{41.70}/\textcolor{red}{33.91} \\
LI  & 36.47/30.22 & 34.38/29.14 & 36.53/29.74 & 34.23/28.89 & 37.26/30.29 & 36.65/30.29 & 35.38/29.39 & 35.74/30.30 & 28.76/22.87 & 37.43/\textcolor{red}{31.35} \\
LR  & 35.92/27.62 & 34.71/26.32 & 35.87/27.14 & 33.63/26.58 & 37.42/28.31 & \textcolor{red}{37.91}/\textcolor{red}{29.46} & 35.04/27.48 & 36.42/28.17 & 30.45/20.74 & \textcolor{blue}{37.55}/\textcolor{blue}{29.26} \\
BC  & 38.08/30.92 & 38.84/29.55 & 40.19/30.93 & 37.44/29.75 & 41.19/31.84 & 40.55/32.14 & 37.87/30.57 & 43.53/\textcolor{blue}{33.86} & 27.17/20.17 & 43.34/33.81 \\
HI  & 31.63/22.51 & 34.30/22.52 & \textcolor{blue}{35.40}/\textcolor{blue}{25.85} & 33.63/23.42 & 32.41/23.42 & \textcolor{red}{40.43}/\textcolor{red}{31.01} & 20.96/17.41 & 32.18/22.64 & 31.30/19.26 & 26.71/20.72 \\
SA  & 37.95/30.81 & 40.22/31.04 & 36.90/30.43 & 39.57/30.78 & \textcolor{blue}{41.40}/32.49 & 36.83/30.74 & 40.82/31.47 & \textcolor{red}{41.67}/\textcolor{red}{33.64} & 29.55/22.39 & 36.81/31.07 \\
FL  & 35.34/25.39 & 28.99/21.41 & 35.52/24.97 & 33.36/24.01 & \textcolor{blue}{37.15}/26.70 & \textcolor{red}{39.47}/\textcolor{red}{28.73} & 31.51/22.73 & 36.51/25.99 & 30.61/18.59 & 35.61/26.25 \\
SO  & 38.37/29.43 & 36.60/27.32 & 38.82/28.98 & 37.84/29.13 & \textcolor{blue}{41.59}/31.01 & 39.39/30.45 & 35.63/27.80 & 40.85/31.16 & 29.89/21.52 & 41.52/\textcolor{red}{31.80} \\
FM  & 38.72/33.09 & 36.14/30.94 & 38.01/31.43 & 36.28/31.14 & 40.59/33.90 & \textcolor{red}{41.52}/\textcolor{blue}{34.91} & 36.34/31.38 & 39.82/34.01 & 28.70/23.30 & \textcolor{blue}{40.90}/\textcolor{red}{35.13} \\
OV  & 15.28/13.49 & 15.94/13.79 & 16.79/13.27 & 15.50/13.06 & 17.21/13.87 & \textcolor{red}{25.47}/\textcolor{red}{19.35} & 14.70/13.86 & 16.79/14.90 & 12.48/10.71 & 17.55/15.29 \\
SV  & 38.06/30.82 & 37.70/29.65 & 38.88/30.53 & 37.22/30.21 & 40.37/32.02 & 41.31/\textcolor{red}{33.46} & 35.97/29.58 & 40.73/32.72 & 30.46/22.71 & \textcolor{blue}{41.48}/\textcolor{red}{33.46} \\
CM  & 41.25/34.92 & 41.23/32.33 & 37.28/31.52 & 39.37/32.69 & 40.44/33.74 & \textcolor{red}{46.91}/\textcolor{red}{39.21} & 37.85/32.15 & 44.13/36.00 & 28.82/23.45 & 42.43/35.60 \\
TC  & 25.78/21.94 & 26.43/20.71 & \textcolor{blue}{33.30}/22.69 & 25.35/20.62 & 28.29/22.38 & 22.14/19.00 & 25.56/21.20 & \textcolor{red}{37.19}/\textcolor{red}{26.58} & 22.13/17.01 & 30.06/\textcolor{blue}{24.62} \\
ARC & 36.92/30.08 & 36.58/29.05 & 37.71/30.09 & 36.48/29.83 & 39.47/31.49 & 40.64/32.93 & 34.65/28.82 & 39.80/32.30 & 29.44/22.18 & \textcolor{blue}{40.68}/\textcolor{red}{33.12} \\
\hline
MVO & 66.82/49.39 & 61.21/45.12 & 66.99/48.60 & 62.90/46.87 & 75.12/53.85 & 61.07/44.87 & 62.22/46.37 & 74.29/54.80 & 44.56/32.22 & \textcolor{blue}{77.61}/\textcolor{red}{57.24} \\
PVO & 37.01/30.91 & 37.08/28.90 & 36.71/29.86 & 35.60/30.38 & 35.76/30.15 & \textcolor{red}{42.90}/\textcolor{red}{35.96} & 35.61/29.91 & \textcolor{blue}{37.91}/\textcolor{blue}{31.40} & 35.87/28.29 & 36.25/30.93 \\
CV  & 29.35/24.51 & 29.90/24.25 & 30.48/24.40 & 29.43/24.42 & 30.63/25.13 & \textcolor{red}{34.61}/\textcolor{red}{28.46} & 28.05/23.88 & 30.52/25.42 & 23.15/16.70 & \textcolor{blue}{31.45}/\textcolor{blue}{26.12} \\
\hline
\end{tabular}
}

\normalsize
\renewcommand{\arraystretch}{1.0}
\end{table*}

\section{Evaluation and analysis}
In this section, we evaluate 20 tracking algorithms proposed in recent years on the DRGBT-1K dataset, including 16 RGBT trackers, 3 modality-missing trackers, and 1 dynamic RGBT tracker.
Specifically, the RGBT tracking methods include OSTrack~\cite{ye2022ostrack}, TBSI~\cite{tbsi}, TATrack~\cite{TATrack}, BAT~\cite{BAT}, PURA~\cite{PURA}, SDSTrack~\cite{SDSTrack}, MMLoRAT~\cite{LoRAT}, CKD~\cite{CKD}, AINet~\cite{ainet}, STTrack~\cite{sttrack}, CAFormer~\cite{CAFormer}, FMTrack~\cite{xue2025fmtrack}, QSTNet~\cite{ding2025quality}, MRTTrack~\cite{MRTTrack}, UATrack~\cite{UATrack}, and GOLA~\cite{GOLA}. The modality-missing tracking methods include IPT~\cite{IPT}, TMKD~\cite{TMKD}, and SCDT~\cite{SCDT}.
The dynamic RGBT tracking method is CMRL~\cite{DRGBT603}, which exploits invariant representations to improve robustness against modality and viewpoint variations.
Next, we first analyze the overall performance of different trackers on the entire dataset, and then provide a more detailed evaluation under different challenge attributes and dataset subsets.

\begin{figure*}[!htb]
\centering
\includegraphics[width=6.8 in]{ 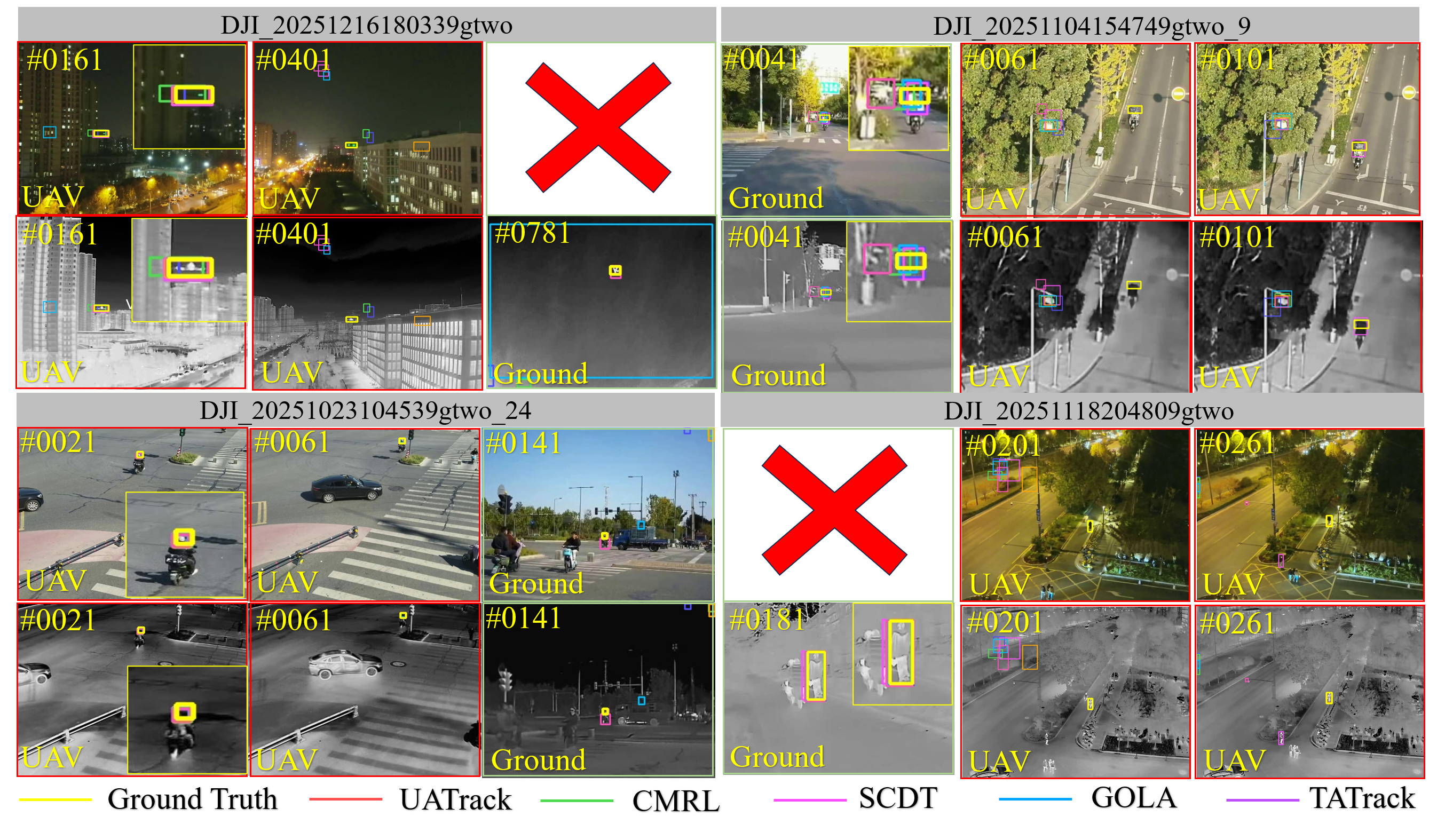}
\caption{Qualitative comparison of state-of-the-art trackers on the DRGBT-1K dataset.
}
\label{vis_results}
\end{figure*}

\subsection{Overall Evaluation Results}
We retrain 20 tracking algorithms on the training set of DRGBT-1K and evaluate them on the test set, with results reported in Table~\ref{tab:sota_comparison}.
Overall, GOLA~\cite{GOLA} consistently achieves the best performance across PR, NPR, and SR, with the highest scores of 51.18\%, 46.42\%, and 38.01\%, respectively. This may be attributed to its ability to effectively suppress the influence of dynamic modality variations, which leads to stronger robustness in complex DRGBT scenarios.
FMTrack~\cite{xue2025fmtrack} achieves competitive results, particularly in PR, where it ranks second with 48.64\%. In terms of NPR, MMLoRAT~\cite{LoRAT} also shows strong performance, reaching 43.06\% and ranking second among all methods.
Compared with RGBT trackers, modality-missing methods such as IPT~\cite{IPT}, TMKD~\cite{TMKD}, and SCDT~\cite{SCDT} generally show inferior performance. This suggests that trackers designed only for modality-missing conditions may struggle to handle the more complex dynamic modality variations in DRGBT tracking scenarios.
Finally, although CMRL~\cite{DRGBT603} is specifically designed for dynamic RGBT tracking scenarios, it mainly focuses on learning invariant representations across modalities and viewpoints, making it difficult to fully exploit modality-specific discriminative cues during dynamic RGBT tracking. This indicates that simply relying on invariant representation learning may weaken the utilization of multimodal complementary information and thus limit tracking performance.

\subsection{Challenge-based Evaluation Results}
To further analyze tracker performance on DRGBT-1K from different perspectives, we evaluate the trackers under 15 challenge attributes and three testing subsets, as reported in Table~\ref{tab:challenge_pr_sr_transposed}.
From the attribute-based results, different trackers show clear advantages under different challenging factors.
For example, GOLA~\cite{GOLA} achieves the best performance under PO in both PR and SR, and also obtains the best SR scores under LI, SO, FM, SV, and ARC.
TATrack~\cite{TATrack} performs well on several appearance- and scale-related challenges, achieving the best PR under BC, SO, SV, and ARC.
CMRL~\cite{DRGBT603} shows strong robustness under several dynamic or degradation-related attributes, such as LR, HI, FL, OV, and CM, where it achieves the best PR and SR scores.
In addition, TMKD~\cite{TMKD} performs competitively under SA and TC, indicating its advantage in handling similar appearance interference and thermal crossover.
Despite these strengths, the results also reveal that several challenges remain difficult for existing trackers.
For instance, the performance on OV and TC is much lower than that on most other attributes, suggesting that target disappearance/reappearance and thermal ambiguity remain severe obstacles in DRGBT tracking.
Moreover, attributes such as LR, SO, SV, and ARC are closely related to viewpoint changes and platform transitions, and they still lead to notable performance degradation for many trackers.
These observations indicate that DRGBT-1K introduces complex real-world challenges that cannot be fully addressed by existing RGBT tracking, modality-missing tracking, or dynamic RGBT tracking methods.

For subset-based evaluation, the results show that trackers achieve substantially higher performance on the MVO subset than on the PVO and CV subsets.
For example, the best PR/SR scores on MVO reach 78.72\%/57.24\%, while the best scores on PVO and CV are only 42.90\%/35.96\% and 34.61\%/28.46\%, respectively.
This indicates that platform variation introduces more severe challenges than modality variation alone, and the combined variation of modality and platform further increases the tracking difficulty.
Notably, CMRL~\cite{DRGBT603} achieves the best performance on both PVO and CV, suggesting that modeling dynamic variations is beneficial for cross-platform scenarios.
However, its overall performance is still limited compared with the best-performing trackers in Table~\ref{tab:sota_comparison}, implying that more effective mechanisms are still needed to jointly exploit modality-specific cues and handle dynamic viewpoint variations.

\subsection{Qualitative Evaluation}
In this section, we provide a qualitative comparison on the DRGBT-1K dataset using five representative state-of-the-art trackers, including UATrack~\cite{UATrack}, CMRL~\cite{DRGBT603}, SCDT~\cite{SCDT}, GOLA~\cite{GOLA}, and TATrack~\cite{TATrack}.
Specifically, four representative sequences are selected from the DRGBT-1K test set for visualization, and the qualitative results are presented in Fig.~\ref{vis_results}.
As shown in Fig.~\ref{vis_results}, these sequences contain several typical challenges in DRGBT-1K, such as low illumination, small targets, scale variation, background clutter, and significant viewpoint changes caused by platform transitions.
Although some trackers can roughly localize the target in relatively stable frames, their predictions tend to drift or deviate from the ground truth when the target undergoes drastic appearance changes across modalities or viewpoints.
For example, in nighttime scenes and UAV-to-ground viewpoint transitions, targets often become small, blurred, or visually ambiguous, making accurate localization difficult.
Moreover, when similar objects or cluttered backgrounds appear around the target, several trackers produce inaccurate bounding boxes or gradually drift toward surrounding distractors.
These qualitative results further demonstrate that DRGBT-1K introduces challenging dynamic RGBT tracking scenarios, requiring trackers to effectively handle multimodal appearance variations and cross-platform viewpoint changes.

\section{Conclusion}
In this work, we present DRGBT-1K, a new large-scale dynamic RGBT tracking benchmark with high-quality annotations.
Meanwhile, we evaluate representative state-of-the-art tracking algorithms proposed in recent years on DRGBT-1K. The experimental analysis shows that existing trackers still face clear limitations in dynamic RGBT tracking scenarios, especially under complex conditions involving both cross-platform viewpoint changes and cross-modal appearance variations. We hope that DRGBT-1K can provide strong data support for object tracking research in cross-platform and cross-modal scenarios, and further promote the development of more robust DRGBT tracking algorithms.
In addition, based on the data resources of DRGBT-1K, we release an unaligned version to facilitate practical applications under real-world sensor misalignment, and construct UGVT-1K, a UAV-ground collaborative tracking dataset, to support multi-platform collaborative visual tracking.
We believe that the release of these high-quality tracking resources will facilitate future research and promote the development of the tracking field.

\bibliography{cite}

@inproceedings{SDSTrack,
  title={Sdstrack: Self-distillation symmetric adapter learning for multi-modal visual object tracking},
  author={Hou, Xiaojun and Xing, Jiazheng and Qian, Yijie and Guo, Yaowei and Xin, Shuo and Chen, Junhao and Tang, Kai and Wang, Mengmeng and Jiang, Zhengkai and Liu, Liang and others},
  booktitle={Proceedings of the IEEE/CVF Conference on Computer Vision and Pattern Recognition},
  pages={26551--26561},
  year={2024}
}

@inproceedings{TATrack,
  title={Temporal Adaptive RGBT Tracking with Modality Prompt},
  author={Wang, Hongyu and Liu, Xiaotao and Li, Yifan and Sun, Meng and Yuan, Dian and Liu, Jing},
  booktitle={Proceedings of the AAAI Conference on Artificial Intelligence},
  year={2024},
  pages={5436--5444}
}

@inproceedings{QAT2023,
  title={Quality-Aware RGBT Tracking via Supervised Reliability Learning and Weighted Residual Guidance},
  author={Liu, Lei and Li, Chenglong and Xiao, Yun and Tang, Jin},
  booktitle={Proceedings of the 31st ACM International Conference on Multimedia},
  pages={3129--3137},
  year={2023}
}

@inproceedings{tbsi,
  title={Bridging Search Region Interaction With Template for RGB-T Tracking},
  author={Hui, Tianrui and Xun, Zizheng and Peng, Fengguang and Huang, Junshi and Wei, Xiaoming and Wei, Xiaolin and Dai, Jiao and Han, Jizhong and Liu, Si},
  booktitle={Proceedings of the IEEE/CVF Conference on Computer Vision and Pattern Recognition},
  pages={13630--13639},
  year={2023}
}

@inproceedings{BAT,
  
  author={Bing Cao and Junliang Guo and Pengfei Zhu and Qinghua Hu},
title={Bi-directional Adapter for Multimodal Tracking},
  booktitle={AAAI Conference on Artificial Intelligence},
pages={927-935},
  year={2024}
}

@inproceedings{sttrack,
       
      author={Xiantao Hu and Ying Tai and Xu Zhao and Chen Zhao and Zhenyu Zhang and Jun Li and Bineng Zhong and Jian Yang},
title={Exploiting Multimodal Spatial-temporal Patterns for Video Object Tracking},
booktitle={Proceedings of the AAAI Conference on Artificial Intelligence, 39(4).},
      year={2024},
      pages={3581-3589}
}

@inproceedings{ye2022ostrack,
author={Ye, Botao and Chang, Hong and Ma, Bingpeng and Shan, Shiguang and Chen, Xilin},
  title={Joint Feature Learning and Relation Modeling for Tracking: A One-Stream Framework},
  pages = {341–357},
  booktitle={ECCV},
  year={2022}
}

@inproceedings{lasher,
  author={Li, Chenglong and Xue, Wanlin and Jia, Yaqing and Qu, Zhichen and Luo, Bin and Tang, Jin and Sun, Dengdi},
title={LasHeR: A Large-Scale High-Diversity Benchmark for RGBT Tracking}, 
  booktitle={IEEE Transactions on Image Processing}, 
  pages={392-404},
  year={2022},

  }

@inproceedings{VTUAV,
author = {Zhang Pengyu and Jie Zhao and Dong Wang and Huchuan Lu and Xiang Ruan},
title = {Visible-Thermal UAV Tracking: A Large-Scale Benchmark and New Baseline},
booktitle = {Proceedings of the IEEE conference on computer vision and pattern recognition},
year = {2022}
}

@inproceedings{gtot,
  author={Li, Chenglong and Cheng, Hui and Hu, Shiyi and Liu, Xiaobai and Tang, Jin and Lin, Liang},
title={Learning Collaborative Sparse Representation for Grayscale-Thermal Tracking}, 
  booktitle={IEEE Transactions on Image Processing}, 
  pages={5743-5756},
  year={2016},

  
  }

@inproceedings{AFTER,
title={AFTER: Attention-Based Fusion Router for RGBT Tracking}, 
  author={Lu, Andong and Wang, Wanyu and Li, Chenglong and Tang, Jin and Luo, Bin},
  booktitle={IEEE Transactions on Image Processing}, 
  
  pages={4386-4401},
year={2025},
}

@inproceedings{CAFormer,
  title={Cross-modulated Attention Transformer for RGBT Tracking},
  author={Yun Xiao and Jiacong Zhao and Andong Lu and Chenglong Li and Bing Yin and Yin Lin and Cong Liu},
  booktitle={AAAI Conference on Artificial Intelligence},
pages={8682-8690},
  year={2025},

}

@inproceedings{ainet,
      title={RGBT Tracking via All-layer Multimodal Interactions with Progressive Fusion Mamba}, 
      author={Andong Lu and Wanyu Wang and Chenglong Li and Jin Tang and Bin Luo},
booktitle={Proceedings of the AAAI Conference on Artificial Intelligence},
pages={5793-5801},
      year={2025},
}

@inproceedings{ding2025quality,
  title={Quality-Aware Spatio-Temporal Transformer Network for RGBT Tracking},
  author={Ding, Zhaodong and Li, Chenglong and Wang, Tao and Wang, Futian},
  booktitle={IEEE Transactions on Image Processing},
  pages={7845--7858},
  year={2025},
}

@inproceedings{DRGBT603,
  title={Causality-based Modality\&Platform-invariant Representation Learning for Dynamic RGBT Tracking and A Benchmark},
  author={Ding, Zhaodong and Li, Chenglong and Miao, Shengqing and Tang, Jin},
  booktitle={IEEE Transactions on Image Processing},
  pages={3141 - 3156},
  year={2026},
}

@inproceedings{rgbt210,
  title={Weighted sparse representation regularized graph learning for RGB-T object tracking},
  author={Li, Chenglong and Zhao, Nan and Lu, Yijuan and Zhu, Chengli and Tang, Jin},
  booktitle={Proceedings of the 25th ACM international conference on Multimedia},
  pages={1856--1864},
  year={2017}
}

@inproceedings{rgbt234,
  title={RGB-T object tracking: Benchmark and baseline},
  author={Li, Chenglong and Liang, Xinyan and Lu, Yijuan and Zhao, Nan and Tang, Jin},
  booktitle={Pattern Recognition},
  volume={96},
  pages={106977},
  year={2019},
  publisher={Elsevier}
}

@inproceedings{catpp,
  title={RGBT Tracking via Challenge-Based Appearance Disentanglement and Interaction},
  author={Liu, Lei and Li, Chenglong and Xiao, Yun and Ruan, Rui and Fan, Minghao},
  booktitle={IEEE Transactions on Image Processing},
  year={2024},
}

@inproceedings{xue2025fmtrack,
  title={FMTrack: Frequency-aware interaction and multi-expert fusion for RGB-T tracking},
  author={Xue, Yuanliang and Jin, Guodong and Zhong, Bineng and Shen, Tao and Tan, Lining and Xue, Chaocan and Zheng, Yaozong},
  booktitle={IEEE Transactions on Circuits and Systems for Video Technology},
  pages={1655--1667},
  year={2025}
}

@inproceedings{ding2025template,
  title={Template-based uncertainty multimodal fusion network for RGBT tracking},
  author={Ding, Zhaodong and Li, Chenglong and Miao, Shengqing and Tang, Jin},
  booktitle={Proceedings of the Thirty-Fourth International Joint Conference on Artificial Intelligence, IJCAI-25},
  pages={909--917},
  year={2025}
}

@inproceedings{feng2024rgbt,
  title={RGBT tracking: A comprehensive review},
  author={Feng, Mingzheng and Su, Jianbo},
  booktitle={Information Fusion},
  pages={102492},
  year={2024}
}

@inproceedings{ragtrack,
  title={RAGTrack: Language-aware RGBT Tracking with Retrieval-Augmented Generation},
  author={Li, Hao and Wang, Yuhao and Hao, Wenning and Zhang, Pingping and Wang, Dong and Lu, Huchuan},
  booktitle={Proceedings of the IEEE/CVF Conference on Computer Vision and Pattern Recognition},
  pages={28179--28189},
  year={2026}
}

@inproceedings{IPT,
  title={Modality-missing RGBT tracking: Invertible prompt learning and high-quality benchmarks},
  author={Lu, Andong and Li, Chenglong and Zhao, Jiacong and Tang, Jin and Luo, Bin},
  booktitle={International Journal of Computer Vision},
  pages={2599--2619},
  year={2025}
}

@inproceedings{huang2024rtracker,
  title={Rtracker: Recoverable tracking via pn tree structured memory},
  author={Huang, Yuqing and Li, Xin and Zhou, Zikun and Wang, Yaowei and He, Zhenyu and Yang, Ming-Hsuan},
  booktitle={Proceedings of the IEEE/CVF Conference on Computer Vision and Pattern Recognition},
  pages={19038--19047},
  year={2024}
}

@inproceedings{trackingnet,
  title={Trackingnet: A large-scale dataset and benchmark for object tracking in the wild},
  author={Muller, Matthias and Bibi, Adel and Giancola, Silvio and Alsubaihi, Salman and Ghanem, Bernard},
  booktitle={Proceedings of the European conference on computer vision (ECCV)},
  pages={300--317},
  year={2018}
}

@inproceedings{SCDT,
  title={Spatio-Temporal Conditional Denoising Transformer for Modality-Missing RGBT Tracking},
  author={Lu, Andong and Zha, Ziyi and Jin, Jiandong and Li, Shihao and Li, Chenglong and Tang, Jin and Luo, Bin},
  booktitle={Proceedings of the IEEE/CVF Conference on Computer Vision and Pattern Recognition},
  pages={13584--13593},
  year={2026}
}

@inproceedings{CKD,
  title={Breaking modality gap in RGBT tracking: Coupled knowledge distillation},
  author={Lu, Andong and Zhao, Jiacong and Li, Chenglong and Xiao, Yun and Luo, Bin},
  booktitle={Proceedings of the 32nd ACM International Conference on Multimedia},
  pages={9291--9300},
  year={2024}
}

@inproceedings{UATrack,
  title={Uncertainty-Aware RGBT Tracking},
  author={Ding, Zhaodong and Li, Chenglong and Wang, Futian and Tang, Jin},
  booktitle={International Journal of Computer Vision},
  pages={290},
  year={2026}
  }

@inproceedings{MRTTrack,
  title={Mining representative tokens via transformer-based multi-modal interaction for RGB-T tracking},
  author={Lai, Pujian and Gao, Dong and Wang, Shilei and Cheng, Gong},
  booktitle={Pattern Recognition},
  pages={112162},
  year={2025},
}

@article{TMKD,
  title={Temporal multimodal knowledge distillation for modality-missing RGBT tracking},
  author={Ruan, Rui and Kang, Yunlong and Liu, Lei and Sun, Jingpeng and Li, Chenglong and Tang, Jin},
  journal={Pattern Recognition},
  pages={113991},
  year={2026},
  publisher={Elsevier}
}

@inproceedings{GOLA,
  title={Group Orthogonal Low-Rank Adaptation for RGB-T Tracking},
  author={Shao, Zekai and Hu, Yufan and Liu, Jingyuan and Fan, Bin and Liu, Hongmin},
  booktitle={Proceedings of the AAAI Conference on Artificial Intelligence},
  volume={40},
  number={11},
  pages={8887--8895},
  year={2026}
}

@inproceedings{PURA,
  title={PURA: Parameter Update-Recovery Test-Time Adapation for RGB-T Tracking},
  author={Shao, Zekai and Hu, Yufan and Fan, Bin and Liu, Hongmin},
  booktitle={Proceedings of the Computer Vision and Pattern Recognition Conference},
  pages={22089--22098},
  year={2025}
}

@inproceedings{LoRAT,
  title={Tracking meets lora: Faster training, larger model, stronger performance},
  author={Lin, Liting and Fan, Heng and Zhang, Zhipeng and Wang, Yaowei and Xu, Yong and Ling, Haibin},
  booktitle={European Conference on Computer Vision},
  pages={300--318},
  year={2024},
}

@inproceedings{jin2026progressive,
    author    = {Jin, Jiandong and Li, Chenglong and Feng, Hao and Lu, Andong and Huang, Lili and Tang, Jin},
    title     = {Progressive Multi-cue Alignment for Unaligned RGBT Tracking},
    booktitle = {Proceedings of the IEEE/CVF Conference on Computer Vision and Pattern Recognition (CVPR)},
    year      = {2026},
    pages     = {35207-35216}
}

@inproceedings{xiao2026unaligned,
  title={Unaligned UAV RGBT Tracking: A Largescale Benchmark and a Novel Approach},
  author={Xiao, Yun and Wang, Yuhang and Jin, Jiandong and Zhang, Wankang and Li and Chenglong},
  booktitle={Proceedings of the AAAI Conference on Artificial Intelligence},
  pages={11014--11022},
  year={2026}
}

@inproceedings{sun2023uav,
  title={UAV-ground visual tracking: A unified dataset and collaborative learning approach},
  author={Sun, Dengdi and Cheng, Leilei and Chen, Song and Li, Chenglong and Xiao, Yun and Luo, Bin},
  booktitle={IEEE Transactions on Circuits and Systems for Video Technology},
  pages={3619--3632},
  year={2023}
}

@article{xu2026vl,
  title={VL-UniTrack: A Unified Framework with Visual-Language Prompts for UAV-Ground Visual Tracking},
  author={Xu, Boyue and Hou, Ruichao and Ren, Tongwei and Wu, Gangshan},
  journal={arXiv preprint arXiv:2605.04574},
  year={2026}
}

\vspace{-1 cm}
\begin{IEEEbiography}[{\includegraphics[width=1in,height=1.25in,clip,keepaspectratio]{ 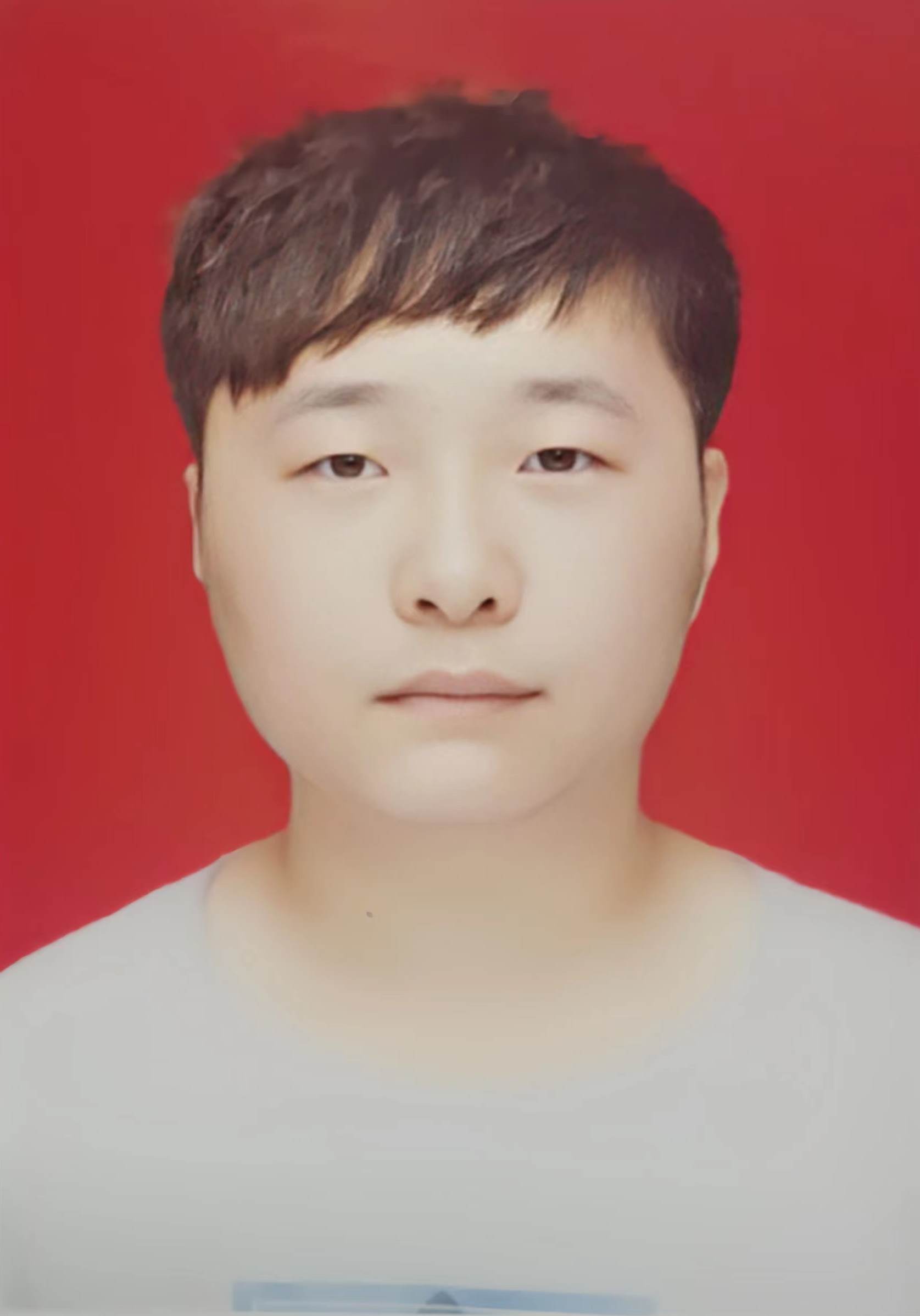}}]{Zhaodong Ding}
is currently pursuing the Ph.D. degree at the School of Artificial Intelligence, Anhui University, China. His research interests include computer vision and deep learning.
\end{IEEEbiography}
\vspace{\biosep}

\begin{IEEEbiography}[{\includegraphics[width=0.9in,height=1.2in,clip,keepaspectratio]{ 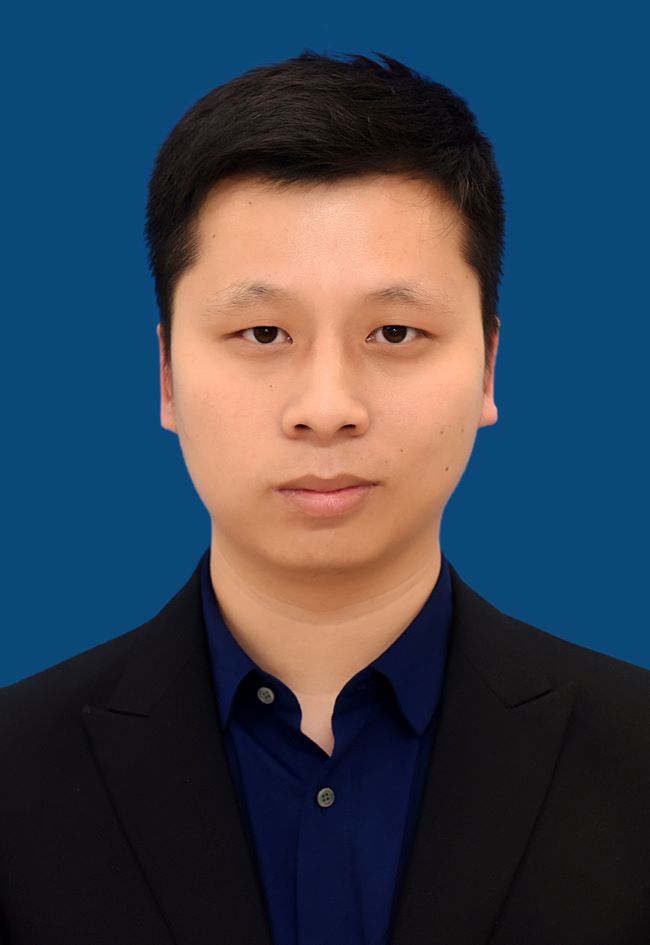}}]{Chenglong Li}
(Senior Member, IEEE) received the M.S. and Ph.D. degrees from the School of Computer Science and Technology, Anhui University, Hefei, China, in 2013 and 2016, respectively. From 2014 to 2015, he was a Visiting Student with the School of Data and Computer Science, Sun Yat-sen University, Guangzhou, China. He was also a Postdoctoral Research Fellow with the Center for Research on Intelligent Perception and Computing (CRIPAC), National Laboratory of Pattern Recognition (NLPR), Institute of Automation, Chinese Academy of Sciences (CASIA), Beijing, China. He is currently a Professor with the State Key Laboratory of Opto-Electronic Information Acquisition and Protection Technology and the School of Artificial Intelligence, Anhui University. His research interests include computer vision and deep learning.
\end{IEEEbiography}
\vspace{\biosep}

\begin{IEEEbiography}[{\includegraphics[width=1in,height=1.25in,clip,keepaspectratio]{ 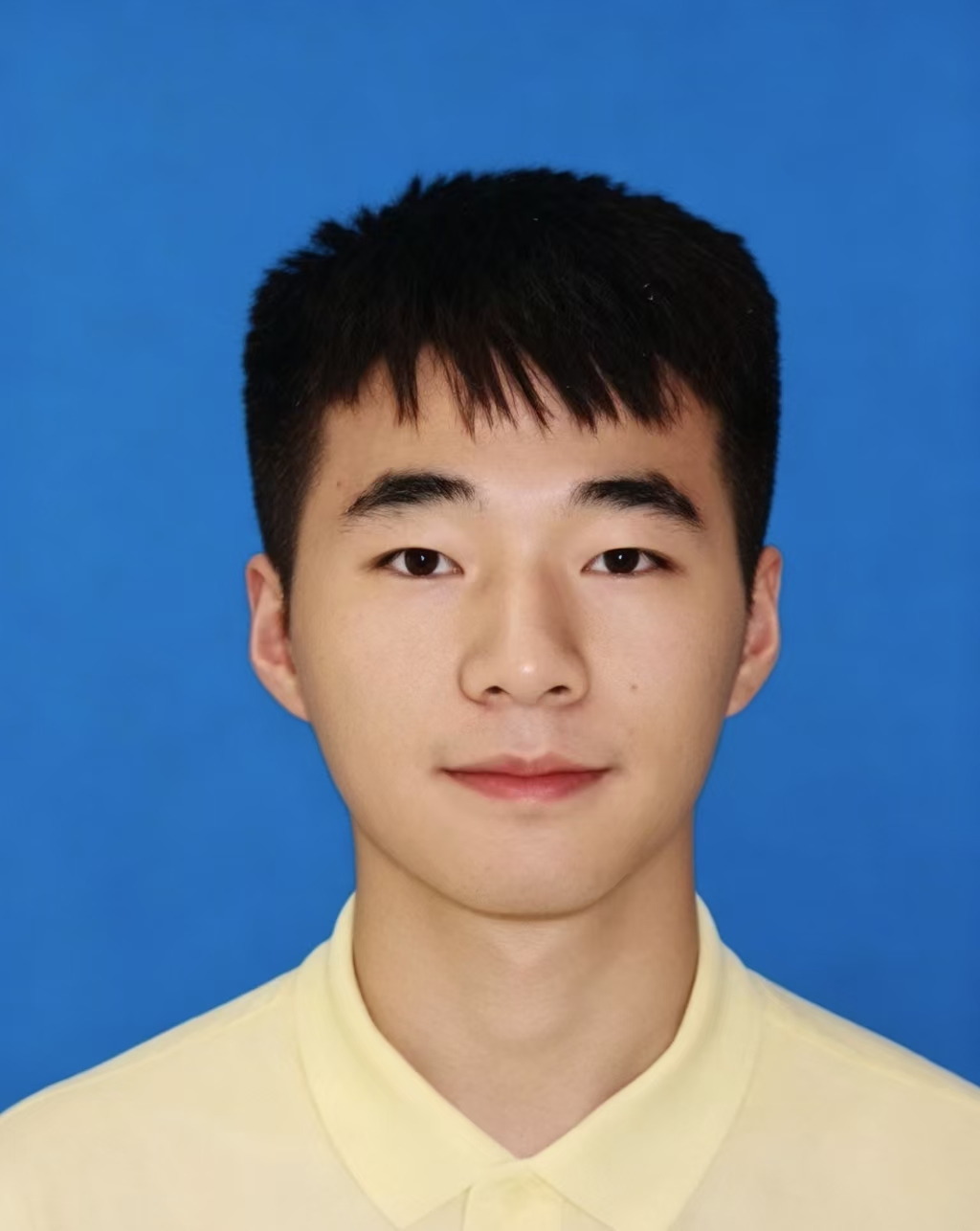}}]{Zeyu Ding}
is currently pursuing the bachelor's degree at the School of Computer Science and Technology, Anhui University, Hefei, China. His research interests include computer vision, deep learning, and multi-modal object tracking.
\end{IEEEbiography}

\begin{IEEEbiography}[{\includegraphics[width=0.9in,height=1.2in,clip,keepaspectratio]{ 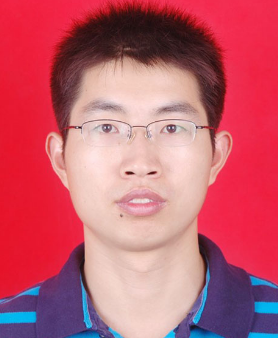}}]{Futian Wang}
received the B.S., M.S., and Ph.D. degrees from the School of Computer Science and Technology, Anhui University, Hefei, China, in 2005, 2009, and 2017, respectively. He is currently an Associate Professor with the School of Computer Science and Technology, Anhui University. His research interests include image processing, computer vision, and edge computing.
\end{IEEEbiography}
\vspace{\biosep}

\begin{IEEEbiography}[{\includegraphics[width=0.95in,height=1.25in,clip,keepaspectratio]{ 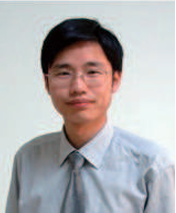}}]{Jin Tang}
received the B.Eng. degree in automation and the Ph.D. degree in computer science from Anhui University, Hefei, China, in 1999 and 2007, respectively. He is currently a Professor with the School of Computer Science and Technology, Anhui University. His current research interests include computer vision, pattern recognition, machine learning, and deep learning.
\end{IEEEbiography}

\end{document}